\documentclass[sigconf]{acmart}
\usepackage{multirow}
\usepackage{float}
\usepackage{enumitem} 
\usepackage{comment}

\usepackage{amssymb}
\usepackage{soul}
\usepackage{multicol}
\AtBeginDocument{%
  }

\setcopyright{none}
\copyrightyear{}
\acmYear{2025}
\acmDOI{}
\acmConference[]{}{}{} 
\acmISBN{}
\begin{document}

\title{TT-LoRA MoE: Unifying Parameter-Efficient Fine-Tuning and Sparse Mixture-of-Experts}

\author{Pradip Kunwar}

\orcid{0009-0004-2583-5925}
\affiliation{%
  \institution{Tennessee Tech University, \\Los Alamos National Laboratory}
  \city{Cookeville}
 \state{Tennessee}
 \country{USA}
}
\email{pkunwar42@tntech.edu}

\author{Minh N. Vu}
\affiliation{%
  \institution{Los Alamos National Laboratory}
  \city{Los Alamos}
 \state{New Mexico}
 \country{USA}
  }
\email{mvu@lanl.gov}

\author{Maanak Gupta}
\affiliation{%
  \institution{Tennessee Tech University}
   \city{Cookeville}
 \state{Tennessee}
 \country{USA}
}
\email{mgupta@tntech.edu}

\author{Mahmoud Abdelsalam}
\affiliation{%
 \institution{Mahmoud Abdelsalam}
 \city{Greensboro}
 \state{North Carolina}
 \country{USA}
 }
\email{mabdelsalam1@ncat.edu}

\author{Manish Bhattarai}
\affiliation{%
  \institution{Los Alamos National Laboratory}
  \city{Los Alamos}
 \state{New Mexico}
 \country{USA}
  }
\email{ceodspspectrum@lanl.gov}

\renewcommand{\shortauthors}{Kunwar et al.}

\begin{abstract}
We propose Tensor-Trained Low-Rank Adaptation Mixture of Experts (TT-LoRA MoE), a novel computational framework integrating Parameter-Efficient Fine-Tuning (PEFT) with sparse MoE routing to address scalability challenges in large model deployments. Unlike traditional MoE approaches, which face substantial computational overhead as expert counts grow, TT-LoRA MoE decomposes training into two distinct, optimized stages. First, we independently train lightweight, tensorized low-rank adapters (TT-LoRA experts), each specialized for specific tasks. Subsequently, these expert adapters remain frozen, eliminating inter-task interference and catastrophic forgetting in multi-task setting. A sparse MoE router, trained separately, dynamically leverages base model representations to select exactly one specialized adapter per input at inference time, automating expert selection without explicit task specification. Comprehensive experiments confirm our architecture retains the memory efficiency of low-rank adapters, seamlessly scales to large expert pools, and achieves robust task-level optimization. This structured decoupling significantly enhances computational efficiency and flexibility: uses only 2\% of LoRA, 0.3\% of Adapters and 0.03\% of AdapterFusion parameters and outperforms AdapterFusion by 4 value in multi-tasking, enabling practical and scalable multi-task inference deployments.
\end{abstract}



\maketitle

\section{Introduction} Large language models (LLMs) have driven remarkable progress in natural language processing (NLP), yet their deployment remains challenging due to the high computational costs and memory demands of full fine-tuning. \emph{Parameter-Efficient Fine-Tuning} (PEFT) approaches—such as LoRA~\cite{hu2021lora} and its tensorized variant TT-LoRA~\cite{anjum2024tensor} mitigate these issues by updating only a small subset of parameters. However, these methods typically require that each task-specific adapter be selected manually during inference, limiting their scalability and practicality in multi-task or dynamic settings.

Concurrently, \emph{Mixture-of-Experts} (MoE) architectures have shown that dynamic routing of inputs to specialized modules can significantly enhance model capacity. Standard MoE systems, though, suffer from drawbacks that include joint expert training overhead, capacity dilution when scaling the number of experts, and difficulties ensuring balanced training across experts.

\begin{table*}[ht]
  \centering
  \caption{Comparison of Recent PEFT-MoE Architectures with Detailed Metadata (2022–2024)}
  \label{tab:summaryofmoe}
  \begin{tabular}{l p{3.1cm} p{3.4cm} p{3.2cm} p{3.2cm}}
    \toprule
    \textbf{Method} & \textbf{Knowledge Forgetting} & \textbf{Inter-task Interference} & \textbf{MoE Scalability} & \textbf{Compute/Memory Efficiency} \\
    \midrule
    AdaMix \cite{wang2022adamix} 
        & --- 
        & --- 
        & Random routing 
        & --\\[1ex]
        \midrule
    MOELoRA \cite{liu2023moelora} 
        & --- 
        & Mitigates Seesawing Effect 
        & Studies Expert Variations 
        & --\\[1ex]
    SiRA \cite{zhu2023sira} 
        & --- 
        & --- 
        & Capacity Limited Experts 
        & --- \\[1ex]
    LoRAMoE \cite{dou2024loramoe} 
        & Prevents World Knowledge Loss 
        & --
        &Dedicated Experts
        & --\\[1ex]
    MoCLE \cite{gou2023mixture} 
        & --- 
        & Handles Task Conflicts via Clustering 
        & Cluster-specific Experts + a General Expert 
        & --- \\[1ex]
        \midrule
    MoRAL \cite{yang2024moral} 
        &Robust Knowledge Retention 
        & --- 
        & --- 
        & ---\\[1ex]
    MoLA \cite{gao2024higher} 
        & --- 
        & --- 
        & 4 Different Expert Combinations
        & Fewer Params than Uniform Allocation\\[1ex]
    MixLoRA \cite{li2024mixlora} 
        & --- 
        & Dynamic Load Balance Mitigate Interference 
        &Designed for Multi Tasks 
        & Saves 40\% Memory, 30\% Latency \\[1ex]
    MoLE \cite{wu2024mixture} 
        & Learns to Combine Pre-trained LoRAs 
        & Keeps LoRA Experts' Identities Distinct 
        & Hierarchical gating 
        & --- \\[1ex]
    AdaMoLE \cite{liu2024adamole} 
        & --- 
        & -- 
        & Dynamic topk 
        & --- \\[1ex]
    MALoRA \cite{wang2024malora} 
        & --- 
        & --- 
        & Shared Subspace of LoRA A to Reduce Overhead 
        & 30-48\% Fewer Params \\[1ex]
    PERFT \cite{liu2024perft} 
        & --- 
        & --- 
        & Framework for Scalable MoE Fine-tuning 
        & ---\\[1ex]
    MoSLD \cite{zhao2024mosld} 
        & Preserves Multi-task Knowledge 
        & --- 
        & LoRA A is Shared, B is Duplicated 
        & Parameter-efficient Design\\
    \midrule
    \textbf{Ours}
    &\textbf{Pre-trained Experts}
    &\textbf{Dynamic and Accurate Expert Adaptation}
    &\textbf{Incrementally Adaptable Router Design}
    &\textbf{2\% of LoRA Parameters in Experts and 0.03\% of Fusion Parameters in Router}\\
    \bottomrule
    
  \end{tabular}
\end{table*}

To address these challenges, we propose \textbf{TT-LoRA MoE}, a novel framework that unifies the parameter efficiency of tensorized low-rank adapters with the dynamic routing benefits of sparse MoE. Our approach proceeds in two distinct stages: 
\begin{enumerate}
\item \textbf{Independent Expert Training:} We train a suite of TT-LoRA adapters independently for each downstream task. Each adapter is highly compressed via tensor-train decomposition, which reduces its parameter count by over 98\% compared to standard LoRA while achieving competitive or superior task performance.
\item \textbf{Dynamic Expert Routing:} Once trained, these TT-LoRA experts are frozen and integrated using a lightweight, noisy top-1 gating router. Leveraging only the base model’s robust hidden representations, the router dynamically selects the appropriate expert for each input in a task-agnostic manner, thereby eliminating the need for manual adapter selection.
\end{enumerate}

Our method has several important advantages over existing approaches. Traditional PEFT systems, despite being parameter-efficient, require manual intervention to choose the right adapter and suffer from scalability issues as the number of tasks grows. In contrast, SOTA MoE methods that jointly train experts and routers are burdened by training instability and increased computational costs. TT-LoRA MoE decouples these concerns by isolating expert learning from routing, which prevents inter-task interference and catastrophic forgetting while enabling efficient and dynamic multi-task adaptation.

The main contributions of our work are: 
\begin{itemize}[leftmargin=*] 
\item A two-stage TT-LoRA MoE framework that decouples expert training from routing, allowing task-specific experts to be learned independently and integrated efficiently. 
\item The introduction of efficient  tensor contraction operation for TT-LoRA adapters, achieving dramatic reductions in trainable parameter counts and faster inference without compromising accuracy. 
\item A lightweight, noise-augmented routing mechanism that leverages base model representations to automatically select experts during inference, thereby obviating manual adapter selection. 
\item Empirical validation that our framework attains state-of-the-art performance across diverse tasks while significantly reducing memory and computational overhead compared to methods such as AdapterFusion and standard LoRA approaches. \end{itemize}


\section{Literature Review and Motivation}
\label{sec:litreview}
In recent years, the explosive growth in the size and capability of LLMs has driven impressive progress in NLP, but this progress brings pressing challenges in efficiency, scalability, and knowledge retention. In this section, we survey  key milestones in model scaling, parameter‑efficient fine‑tuning, and sparse Mixture‑of‑Experts (MoE) from the Multi-Task Learning(MTL) perspective, and we motivate our approach.

\textbf{Scaling LLMs:}
The transformer architecture \cite{vaswani2017attention} inaugurated the modern NLP era by introducing self‑attention and parallelism. Early models such as GPT‑1 \cite{radford2018improving} demonstrated that unsupervised pre‑training followed by supervised fine‑tuning could outperform task‑specific networks, while GPT‑2 \cite{radford2019language} scaled up parameters by 10‑fold and exhibited multi‑task capabilities. Kaplan et al. \cite{kaplan2020scaling} formalized these trends with scaling laws showing predictable improvements with simultaneous increases in model size and data. However, as models such as the 175B GPT‑3 \cite{brown2020language} demonstrate, this scaling comes at significant training and inference costs, which in turn multiply the expense of adapting such models to specific tasks. And this hinders full scale multi-task fine-tuning effectiveness as it's expensive as well as this comes with the risk of loosing base knowledge.

\textbf{Rise of PEFTs:}
Full fine‑tuning of multi‑billion‑parameter LLMs is prohibitively expensive, motivating PEFT strategies that freeze the backbone and learn lightweight, task‑specific components. Methods such as \textbf{Prefix Tuning} \cite{li2021prefix} and \textbf{Prompt Tuning} \cite{lester2021power} introduce virtual tokens or input embeddings to achieve competitive accuracy with minimal parameters. Adapter modules \cite{houlsby2019parameter} only update around 3.6\% of BERT’s weights, while \textbf{LoRA} \cite{hu2021lora} learns rank‑decomposed weight updates, usually adding less than 1\% parameters as compared to the base model parameters count. More recent variants like \textbf{AdaLoRA} \cite{zhang2023adalora} and \textbf{QLoRA} \cite{dettmers2023qlora} further refine the efficiency. In particular, \textbf{TT‑LoRA} \cite{anjum2024tensor} decomposes LoRA weight updates into tensor‑train cores, cutting trainable parameters and memory footprint by one to two orders of magnitude while matching—or even slightly exceeding—the performance of standard LoRA. However, these PEFT techniques inherently require an independent lightweight module for each task, making them less scalable in multi‑task and continual‑learning scenarios.

\textbf{From Single‑Task PEFT to Multi‑Task Modularity:} 
Multi-task learning is promising for general performance improvement, but integrating a new task typically introduces challenges such as catastrophic forgetting \cite{french1999catastrophic, mccloskey1989catastrophic, phang2018sentence, pruksachatkun2020intermediate} and inter‑task interference. Early approaches like \textbf{Cross‑Stitch Networks} \cite{misra2016cross} and \textbf{Deep Adaptation Modules} \cite{rosenfeld2018incremental} share features across tasks, yet they face scalability limitations and often require manual routing. Approaches like \textbf{AdapterHub} \cite{pfeiffer2020adapterhub} and \textbf{AdapterFusion} \cite{pfeiffer2020adapterfusion} mitigate forgetting by dynamically combining pre‑trained adapters, but the fusion layers add significant extra parameters and incur additional inference latency. Recently, \textbf{LoRAHub} \cite{huang2023lorahub} explores keeping a set of LoRA experts frozen and learning a task‑agnostic combination, although the routing remains static and inefficient when tasks grow. These limitations motivate the need for a more scalable and dynamic mechanism to integrate parameter-efficient modules across tasks—leading to the emergence of hybrid approaches that combine PEFT with Mixture-of-Experts architectures.

\textbf{Sparse Mixture‑of‑Experts (MoE) Meets PEFT:} 
Several recent works have attempted to blend Sparse MoE architectures with PEFT methods to address these challenges (see Table~\ref{tab:summaryofmoe}). These hybrid models aim to leverage the modularity and routing flexibility of MoE while retaining the parameter efficiency of PEFT. Yet, existing PEFT‑MoE systems have three primary bottlenecks:

\begin{enumerate}
  \item \textbf{Parameter and Memory Footprint}: Each expert maintains its own complete set of PEFT weights, resulting in a linear increase in parameters as new tasks are added. 
    \item \textbf{Training Overhead}: Joint training of multiple experts and the corresponding routing mechanisms increases computational cost and can lead to instability, especially as the number of tasks grows. 
  \item \textbf{Scalability and Incremental Learning}: Integrating a new task typically requires retraining a large portion of the model, which hinders efficient incremental learning. 
 \end{enumerate}

Our approach, \textbf{TT-LoRA MoE}, distinctly targets these challenges by:
(i) compressing attention weights (Query, Value) using tensor-train decomposition—thereby shrinking storage by an order of magnitude;
(ii) employing a sparse-instance-level router that \textbf{leverages the base model's knowledge} to enhance routing capability, reduce parameter count and training overhead for effective routing; and (iii) supporting incremental learning by \emph{adaptively allocating experts} for transformer layers based on the task type detected by the router.
This comprehensive strategy significantly enhances model scalability and efficiency, setting our methodology apart from existing PEFT-MoE architectures.

\begin{figure}[ht]
  \centering
  \includegraphics[width=\linewidth]{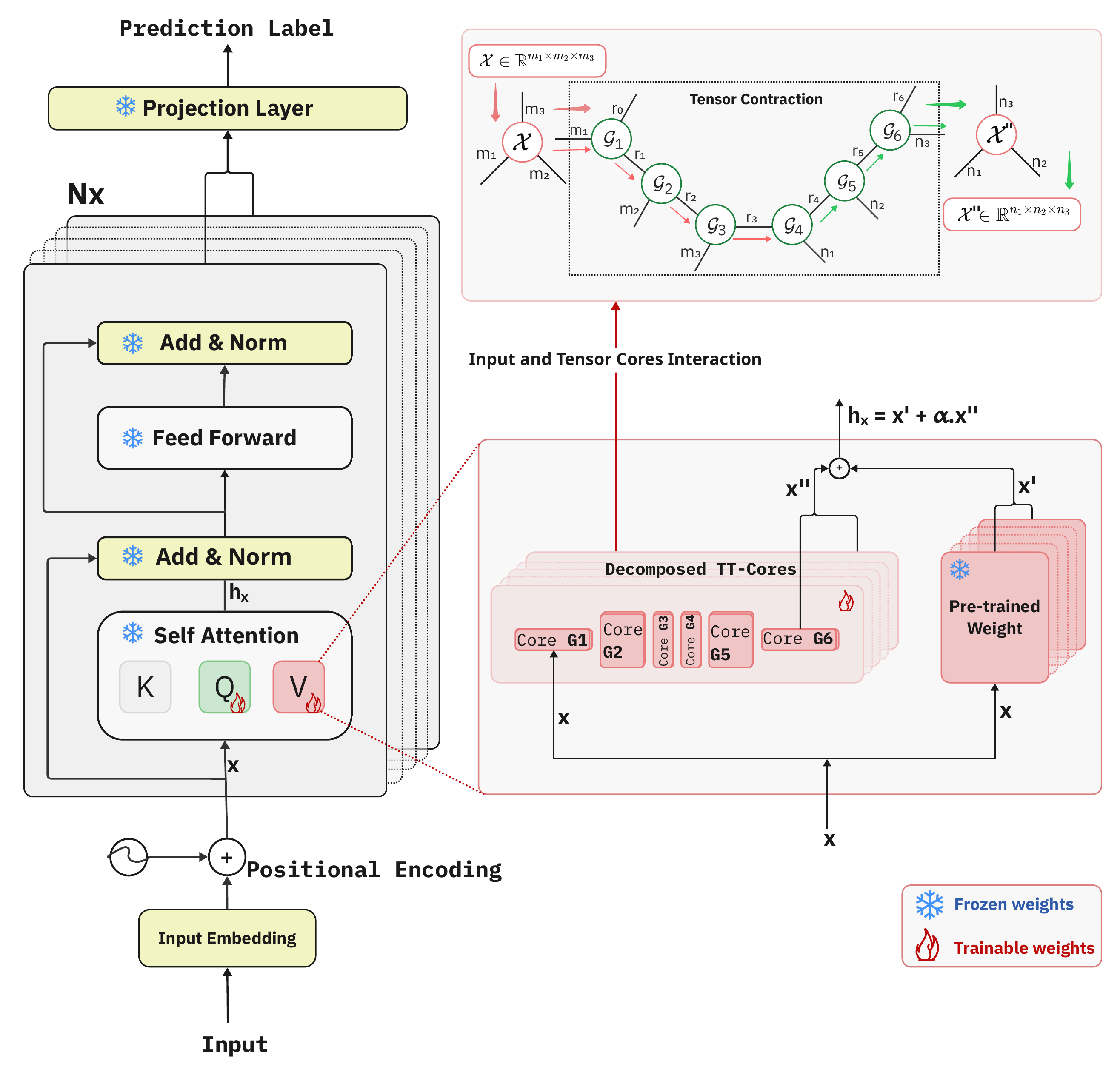}
  \caption{Training Single TT-LoRA Expert}
  \label{fig:expert-training}
  \Description{Expert training in individual dataset and the cores along with projection layer is saved for the that task}
\end{figure}

\section{Preliminaries}
\label{sec:background}
In this section, we briefly introduce the essential background components that underlie our proposed TT-LoRA MoE framework.

\textbf{PEFT Approaches:} 
Parameter-Efficient Fine-Tuning methods are designed to adapt large pre-trained language models to downstream tasks while updating only a small fraction of model parameters and reduces the overhead of adapting large pretrained models. PEFT strategies freeze the backbone model weights and introduce lightweight trainable components, significantly reducing the computational cost and storage overhead associated with full fine-tuning. 

Instead of updating the full model parameters $\Theta$, adapters\cite{houlsby2019parameter, pfeiffer2020adapterhub} introduce a small set of learnable weights $\Phi$, keeping the original backbone frozen. Each adapter consists of a bottleneck feed-forward layer with a down-projection weight matrices, a non-linearity activation, and an up-projection weight matrix. The output is given by $Adapter(h)=h+W_{up}.f(W_{down}.h)$, where $h\in \mathbb{R}^{d}$, $W_{down}\in\mathbb{R}^{r\times d}$ and $W_{up}\in\mathbb{R}^{d\times r}$. Similarly in LoRA~\cite{hu2021lora}, the full weight matrix $W_0\in \mathbb{R}^{m\times n}$ is not updated directly. Instead, LoRA inserts a trainable update in the form $\Delta W = BA$, where $A\in\mathbb{R}^{r\times n}$, $B\in\mathbb{R}^{m\times r}$, and $r\ll min(m,n)$. The layer output then becomes $h=W_0x + \alpha.BAx$ where only $B$ and $A$ are trainable. Extending this principle, TT-LoRA~\cite{anjum2024tensor} decomposes the adapter into a chain of small “3D tensor” cores. For an input vector $x\in \mathbb{R}^{d_{in}}$, the weight vector $W\in\mathbb{R}^{d_{in}\times d_{out}}$ is reshaped into a tensor $\mathcal{W} \in \mathbb{R}^{m_1 \times m_2 \times \dots \times m_p \times n_1 \times n_2 \dots \times n_q}$ where $d_{in} = m_1 \times m_2 \times \dots \times m_p$, $d_{out} = n_1 \times n_2 \dots \times n_q$ and a series of TT-cores $\mathcal{G}=\{G^{(1)},G^{(2)},...,G^{(p+q)}$\} is defined, where each core $G^{(i)}\in \mathbb{R}^{r_{i-1}\times (m_i\ or\ n_i) \times r_i}$ where $r_i=[1,r,...r,1]$. Similar to LoRA and other PEFT methods, these low rank representations are typically reconstructed to form the weight matrix at  run-time as $W_0+\alpha. \Delta W$, where $\Delta W$ augments $W$ with the new knowledge from the fine-tuned task. In contrast, our TT-LoRA implementation employs a tensor contraction strategy: the input undergoes series of contraction to each tensor core, thereby obviating the need to reconstruct ${W_0}$ at runtime (more details are provided in Section \ref{subsec:tt-forward}). In addition to reducing the number of trainable parameters, the PEFT approaches requires a separate training of each domain expert (fine-tuning) and multiple trainings for multiple domain experts. However, in the multitask setting, the user must \emph{manually} select the appropriate PEFT to apply and to perform any specific task domain. 

\begin{figure*}[t]
  \centering
  \includegraphics[width=\linewidth]{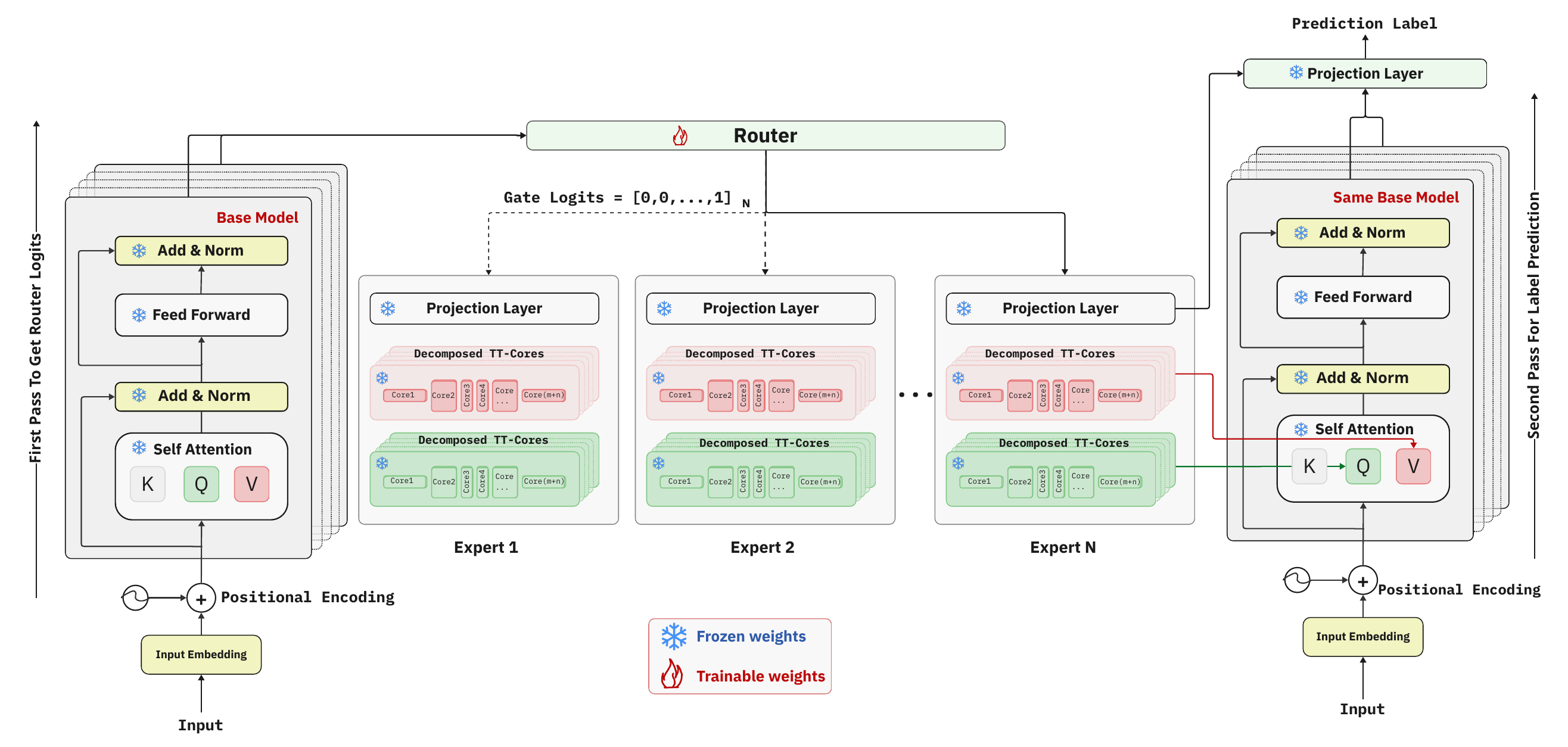}
  \caption{TT-LoRA MoE Architecture Diagram}
  \label{fig:moearchitecture}
  \Description{Both the models used here are the same base model, they are shown different for brevity}
\end{figure*}

\textbf{Mixture-of-Experts.}
Mixture-of-Experts (MoE)~\cite{jacobs1991adaptive, jordan1994hierarchical, shazeer2017outrageously} is a conditional computation paradigm that increases model capacity without proportional rise in inference cost.
MoE introduces multiple parallel expert modules $E$  along with a gating (router) mechanism that selects a sparse subset of these experts for each input. For an input hidden vector $h\in \mathbb{R}^{d}$, an MoE layer contains $N$ experts $\{ E_i(\cdot) \}_{i=1}^N$, where each expert is typically a feed-forward network. A gating (learnable router) network $G(h)\in \mathbb{R}^{N}$ assigns a score to each expert as $G(h) = \text{Softmax}(W_g h)$, where $W_g \in \mathbb{R}^{N \times d}$ and is the router's weight matrix. In ${top\mathbf{k}}$ sparse MoE setting (usually $k$=1 or $k$=2), the output is computed as a weighted sum over the top-$k$ selected experts as $\text{MoE}(h) = \sum\limits_{i=1}^N G(h)_i \cdot E_i(h)$, where $G(h)_i$ is the normalized gate score for expert $i$ and $E_i(h)$ is the output of the $i$-th expert. 

A key challenge with standard MoE is that the router encounters issues such as \emph{router collapse} or \emph{expert imbalance} ~\cite{shazeer2017outrageously}, which are typically mitigated by introducing auxiliary loss (Eg., a load balancing loss). However, standard MoE requires \emph{joint} training of the experts with the router, a process that can be computationally expensive and susceptible to “capacity dilution” (a mismatch between theoretical model capacity and effective learning) when the number of experts $E$ is large. In theory, more experts translated to greater capacity and better performance; yet in practice, each input is routed to only one or two experts. As the  expert count $N$ grows, each expert sees fewer inputs, leading to sparse and noisy per-expert training signals, under-trained experts, and a diminished benefit from the increased capacity \cite{fedus2022switch, du2022glam, shazeer2017outrageously}.

We propose bridging these paradigms by treating each PEFT adapter (especially TT-LoRA) as an \emph{expert}, trained \emph{independently} and stored in memory as small low-rank or tensorized structures. We then incorporate a single, small classification layer as router, that leverages the base model knowledge and is trained to automatically select among these adapters at inference time. The key advantages are: i) The router efficiently handles multiple domains or task selection for each input. ii) We can maintain $E$ distinct adapters, each specialized for a particular domain, without incurring joint training overhead. iii)  TT-LoRA ensures that each adapter remains compact, and gating only one adapter per sample, we  prevent a significant increase in runtime costs.

\section{Proposed Architecture and Methodology}
\label{sec:methods}

We propose a two-stage approach to building a Mixture-of-Experts (MoE) from \emph{Tensor-Train LoRA} (\textbf{TT-LoRA}) adapters.  In Stage 1, individual TT-LoRA adapters are trained for each task independently as shown in Fig. ~\ref{fig:expert-training}. In Stage 2, a lightweight router is introduced to dynamically select the appropriate frozen expert during inference as shown in Fig. ~\ref{fig:moearchitecture}. To clearly present our method, we systematically describe each component as: Section~\ref{subsec:problem-statement} defines the problem statement and learning objective functions, Section~\ref{subsec:routing-base-model-leverage} describes the router logit generation method, Section~\ref{subsec:tt-forward} details the precise dimension factorization and forward pass via tensor contraction, Section~\ref{subsec:tt-lora-train} explains the TT-LoRA adapter training, and Section~\ref{subsec:router-train} details training of the sparse router.

\subsection{Problem Statement}
\label{subsec:problem-statement}
In this subsection, we formally define the problem of task-specific adaptation in a multi-task setting. By training an independent TT-LoRA adapter for each task while keeping the base model frozen, we directly address the issue of catastrophic forgetting and inter-task interference. This formulation enables the subsequent routing mechanism to leverage highly specialized, low-parameter experts, thereby ensuring both efficiency and scalability.

Let $\mathcal{T} = \{ \mathcal{D}_1, \dots, \mathcal{D}_N \}$ denote $N$ tasks, where each task $\mathcal{D}_i$ consists of input-label pairs $(x, y) \sim \mathcal{D}_i$. For each task $\mathcal{D}_i$, we train a task-specific TT-LoRA adapter $\phi_i$ while keeping the base model parameters $\Theta$ frozen as shown in Fig. ~\ref{fig:expert-training}. The optimization objective is defined as:

\[
\phi_i^* = \arg\min_{\phi_i} \; \mathbb{E}_{(x, y) \sim \mathcal{D}_i} \; \mathcal{L}(f_{\Theta, \phi_i}(x), y),
\]

where:
\begin{itemize}[label=\textbullet]
  \item $f_{\Theta, \phi_i}(x)$ denotes the output of the base model $\Theta$ augmented with the TT-LoRA adapter $\phi_i$,
  \item $\mathcal{L}(\cdot, \cdot)$ is the task-specific loss function,
  \item $(x, y)$ represents an input and its corresponding ground-truth label from task $\mathcal{D}_i$.
\end{itemize}

After training, all expert adapters $\{ \phi_1, \dots, \phi_N \}$ are frozen and stored for downstream routing and inference. In practice, we adapt TT-LoRA adaption to the Query($Q)$ and Value ($V$) weights across  every layer of the base model as in Fig. ~\ref{fig:expert-training}. Additionally, for each expert adapter, we  store the corresponding classification head (i.e., the projection or scoring layer weights) as shown in Fig. ~\ref{fig:moearchitecture}. During expert training, all base model parameters remain frozen including the classification layer-though these weights are uniformly initialized each time during the model loading, and  the tensor train decomposed randomly initialized copies of $Q$ and $V$ weights' shapes. This setup limits  the number of trainable parameters and ensures efficiency. To enable coordinated and consistent adaptation during routing training, we preserve the trained tensor cores of $Q$ and $V$, as well as the frozen classification head weights, allowing them to be reused in the routing phase without retraining from scratch.

\subsection{Routing via Base Model Representations}
\label{subsec:routing-base-model-leverage}
This subsection introduces the routing mechanism, which is critical for achieving dynamic expert selection. Rather than relying on manual task-specific adapter selection, our routing process leverages the base model’s hidden representations to automatically select the most appropriate TT-LoRA expert as shown in Fig. ~\ref{fig:moearchitecture}. By employing a noisy top-1 gating mechanism, we ensure sparse and deterministic routing, effectively tackling inter-task interference and enhancing overall MoE scalability.

Once all $N$ task-specific experts $\{ \phi_1, \dots, \phi_N \}$ have been trained independently and frozen, we introduce a lightweight routing mechanism to dynamically select the appropriate expert during inference. To encourage sparsity and mimic one-hot expert selection behavior, we follow the noisy top-1 gating mechanism proposed by Shazeer et al. ~\cite{shazeer2017outrageously}. The router comprises  two learned projection matrices: $W_{\text{gate}} \in \mathbb{R}^{d \times N}$ and $W_{\text{noise}} \in \mathbb{R}^{d \times N}$. These projections operate on the hidden representation $h_x \in \mathbb{R}^d$ obtained from the frozen base model $\Theta$ (just before the final projection layer).
The noisy routing score for expert $i$ is computed as:
\[
g_i = (h_x W_{\text{gate}})_i + \mathcal{N}(0, 1) \cdot \textit{Softplus}((h_x W_{\text{noise}})_i),
\]

where:
\begin{itemize}[label=\textbullet]
  \item $(h_x W_{\text{gate}})_i$ is the clean routing score for expert $i$,
  \item $(h_x W_{\text{noise}})_i$ defines the scale of input-dependent noise,
  \item $\mathcal{N}(0, 1)$ is standard normal random variable sampled per expert,
\end{itemize}

Subsequently, we compute the softmax over the noisy logits keeping only the top-$k$ (typically $k$=1) elements to obtain one hot code encoding logits:

\[
\ell = \textit{Softmax }(\textit{topk }(g,k)),
\]
where
\[
    \textit{topk}(v, k)_i = 
    \begin{cases}
        v_i & \textit{if } v_i \textit{ is in the top } k \textit{ elements of } v, \\
        -\infty & \textit{otherwise.}
    \end{cases} 
\]

The selected expert $\phi_z$ (with it's corresponding $Q$ and $V$ tensor cores along with the expert's projection (or score) layer weights are then used to generate the task prediction:

\[
\hat{y} = f_{\Theta, \phi_z}(x)
\]

\paragraph{\textbf{Router Supervision:}}
During training, we supervise the router with the expert label $t \in \{1, \dots, N\}$ that corresponds to the task from which the input $x$ originates. The router loss $\mathcal{L}_{\text{router}}$ is defined using the cross-entropy loss. The final combined loss which combines task prediction loss and router loss is given as:

\[
\mathcal{L}_{\text{total}} = \mathcal{L}_{\text{task}} + \lambda \cdot \mathcal{L}_{\text{router}},
\]

where $\lambda$ is a tunable scalar that balances routing supervision with task performance. At inference time, routing decision becomes deterministic and selects the expert with the highest score, eliminating the need for explicit task identifiers.

\subsection{Tensor-Train Factorization Forward Pass}
\label{subsec:tt-forward}
This block of the section details the tensor-train (TT) factorization used to compress expert parameters. The purpose here is to replace full weight matrices with a series of smaller TT-cores, enabling efficient storage and inference. This factorization is critical for reducing the memory footprint and computational costs—two key gains highlighted in the introduction.


\textit{\textbf{Factorizing} \(\mathbf{W}\):} Consider a weight matrix 
\(\mathbf{W}\in\mathbb{R}^{m\times n}\), we split
\[
  m = m_1 \times m_2 \times \cdots \times m_p,
  \quad
  n = n_1 \times n_2 \times \cdots \times n_q,
\]
so that \(m = \prod_{i=1}^p m_i\) and \(n = \prod_{j=1}^q n_j\). Conceptually, \(\mathbf{W}\) is interpreted as a \((p+q)\)-way tensor of shape 
\(\bigl[m_1,\dots,m_p,\,n_1,\dots,n_q\bigr]\). A \emph{Tensor-Train} (TT) decomposition approximates this \((p+q)\)-dimensional array with a chain of TT-cores 
\[
  \{\mathbf{G}_k\}_{k=1}^{p+q}, 
  \quad
  \mathbf{G}_k\in \mathbb{R}^{\,r_{k-1}\times\mathrm{factorDim}_k\times r_k},
  \quad
  r_0 = r_{p+q}=1,
\]
where each \(\mathrm{factorDim}_k\) corresponds one of the \(m_i\) or \(n_j\). 
\\
\\
\textit{\textbf{Input Reshaping:}} Let \(\mathbf{X}\in \mathbb{R}^{B\times m}\) be a batch of \(B\) inputs. Each row of \(\mathbf{X}\) is reshaped into a \(p\)-way tensor matching \([m_1,\dots,m_p]\). Formally,
\[
  \mathbf{X}_{\text{tensor}}
  \;\in\;
  \mathbb{R}^{\,B,\,m_1,\dots,m_p},
\]
so the \(b\)-th row of \(\mathbf{X}\) becomes a \(p\)-dimensional slice in \(\mathbf{X}_{\text{tensor}}\).
\\
\\
\textit{\textbf{Contracting Input Factors:}} As shown in the upper right part of the Fig. ~\ref{fig:expert-training} about how the input and tensor contraction is performed, this section formalizes how it's performed. Define an initial state \(\mathbf{S}^{(0)}=\mathbf{X}_{\text{tensor}}\in\mathbb{R}^{B\times m_1\times m_2\dots m_p}\). For each \(k=1,\dots,p\), we have corresponding TT-core \(\mathbf{G}_k\in\mathbb{R}^{r_{k-1},m_k,r_k}\). We now define the contracted state $\mathbf{S}^{(k)}$ by performing following contraction
\[
\begin{split}
S^{(k)}(b, r_k, i_{k+1}, \dots, i_p) ={} & \sum_{r_{k-1}=1}^{r_{k-1}} \sum_{i_k=1}^{m_k} S^{(k-1)}(b, r_{k-1}, i_k, i_{k+1}, \dots, i_p) \\
& \quad \times G_k(r_{k-1}, i_k, r_k).
\end{split}
\]

for all \(b \in \{1,\dots,B\}\), \(i_k \in \{1,\dots,m_k\}\), and where \(i_{k+1},\dots,i_p\) denote the indices of any remaining (un-contracted) dimensions.

In other words, prior to the contraction we have:
\[
\mathbf{S}^{(k-1)} \in \mathbb{R}^{B \times r_{k-1} \times m_k \times m_{k+1} \times \cdots \times m_p},
\]
and after contracting over the \(m_k\) dimension with \(\mathbf{G}_k\), we obtain:
\[
\mathbf{S}^{(k)} \in \mathbb{R}^{B \times r_k \times m_{k+1} \times \cdots \times m_p}.
\]

After performing all \(p\) contractions, i.e., after processing all the input dimensions \(m_1,\dots,m_p\), the final state is
\[
\mathbf{S}^{(p)} \in \mathbb{R}^{B \times r_p},
\]
assuming that the contractions eliminate all the original \(m_i\) dimensions.

\textit{\textbf{Appending Output Factors:}}

We now incorporate the output dimensions \(n_1,\dots,n_q\) using the final \(q\) TT-cores. For each \(i = 1,\dots,q\), let the TT-core \(\mathbf{G}_{p+i}\) be defined as
\[
\mathbf{G}_{p+i} \in \mathbb{R}^{r_{p+i-1} \times n_i \times r_{p+i}},
\]

Assume that, at the beginning of the \(i\)th step, the state tensor has the form
\[
\mathbf{S}^{(p+i-1)} \in \mathbb{R}^{B \times r_{p+i-1} \times n_1 \times \cdots \times n_{i-1}},
\]
where for \(i=1\) we interpret \(\mathbf{S}^{(p)} \in \mathbb{R}^{B \times r_p}\) (with no appended output indices). Then, for each \(i = 1,\dots,q\), we define the updated state \(\mathbf{S}^{(p+i)}\) as follows:
\[
\begin{split}
S^{(p+i)}(b, r_{p+i}, j_1, \dots, j_i) ={} & \sum_{r_{p+i-1}=1}^{r_{p+i-1}} S^{(p+i-1)}(b, r_{p+i-1}, j_1, \dots, j_{i-1}) \\
& \quad \times \, G_{p+i}(r_{p+i-1}, j_i, r_{p+i}),
\end{split}
\]
for all \(b \in \{1,\dots,B\}\) and \(j_i \in \{1,\dots,n_i\}\). Here, the indices \(j_1,\dots,j_{i-1}\) come from the previous output dimensions that have already been appended. Consequently, after this contraction, we have
\[
\mathbf{S}^{(p+i)} \in \mathbb{R}^{B \times r_{p+i} \times n_1 \times \cdots \times n_i}.
\]

After processing all \(q\) output factors (i.e., for \(i = q\)), the final state becomes
\[
\mathbf{S}^{(p+q)} \in \mathbb{R}^{B \times r_{p+q} \times n_1 \times \cdots \times n_q}.
\]
Since \(r_{p+q} = 1\), we can write
\[
\mathbf{S}^{(p+q)} \in \mathbb{R}^{B \times 1 \times n_1 \times \cdots \times n_q}.
\]
Flattening the dimensions corresponding to \([n_1, \dots, n_q]\) yields the final output
\[
\mathbf{Y} \in \mathbb{R}^{B \times n},
\]
which approximates \(\mathbf{X}\,\mathbf{W}^{\top}\) (or \(\mathbf{W}\,\mathbf{X}^{\top}\)). Notably, no intermediate step requires forming or storing the full \((m \times n)\) matrix, thereby preserving the efficiency provided by the TT-decomposition.


\subsection{Stage 1: TT-LoRA Adapter Training}
\label{subsec:tt-lora-train}

In this subsection, we describe the training of individual TT-LoRA adapters for each domain. By training experts independently—with the base model frozen—this stage effectively isolates task-specific adaptation, thereby avoiding inter-task interference and catastrophic forgetting. The resulting adapters are compact and efficient, setting the stage for a scalable and modular Mixture-of-Experts architecture.

We now apply this TT factorization to a LoRA-like finetuning scenario, where the base model is frozen. For a given domain \(d\), we define a set of trainable TT-cores \(\{\mathbf{G}_k^{(d)}\}_{k=1}^{p+q}\), each with the same shapes as described previously. If \(\mathcal{D}_d\!=\!\{(\mathbf{x}_s^{(d)}, \mathbf{z}_s^{(d)})\}\) is a labeled dataset for domain \(d\), then the forward pass is modified to add the TT-LoRA update to the base model output:
\[
  \mathbf{y}_s^{(d)}
  \;=\;
  \mathrm{BaseModel}\bigl(\mathbf{x}_s^{(d)}\bigr)
  \;+\;
  \mathrm{TT\_Contract}\!\Bigl(
    \mathbf{x}_s^{(d)},\{\mathbf{G}_k^{(d)}\}
  \Bigr).
\]
The task-specific loss  for domain $d$ is defined as \(\mathcal{L}_d\!\bigl(\mathbf{y}_s^{(d)},\mathbf{z}_s^{(d)}\bigr)\), and the learnable parameters are exactly \(\{\mathbf{G}_k^{(d)}\}\). By minimizing the sum of losses over \(\mathcal{D}_d\), we train an adapter specialized for domain \(d\). After convergence, the  TT-Cores  \(\{\mathbf{G}_k^{(d)}\}\) are \emph{frozen}, thereby producing one TT-LoRA “expert.”

\subsection{Stage 2: Sparse Router on Frozen Experts}
\label{subsec:router-train}
As shown in Fig. ~\ref{fig:moearchitecture}, we describe the integration of individual TT-LoRA experts through a lightweight routing mechanism. By training the router on frozen experts, we enable dynamic, task-agnostic expert selection at inference time—addressing the challenge of manual adapter selection and improving overall scalability. This design also prevents interference across experts by ensuring that each specialist remains unchanged during routing training.

With $E$ TT-LoRA experts $\{\mathbf{G}_k^{(1)}\},\dots,\{\mathbf{G}_k^{(E)}\}$ available, we introduce router function 
\(\mathbf{r}:\mathbb{R}^m \to \mathbb{R}^E\) as described in section ~\ref{subsec:routing-base-model-leverage}. For each input $\mathbf{x}_b$, we compute the router logits $\ell_b=\mathbf{r}(\mathbf{x}_b)$ and yield a one-hot gate vector
\(
  \mathbf{g}_b\in \{0,1\}^E
\)
satisfying $\sum_{e=1}^E g_{b,e}=1.$ 
Let $e^*=\arg\max(\mathbf{g}_b)$; then the final output is  computed as
\[
  \mathbf{y}_b
  \;=\;
  \mathrm{TT\_Contract}\!\Bigl(
    \mathbf{x}_b,\{\mathbf{G}_k^{(e^*)}\}
  \Bigr),
\]
where the TT-cores $\{\mathbf{G}_k^{(d)}\}$ are now \emph{frozen} for all $d$. We only update $W_{gate}$ and $W_{noise}$ in $\mathbf{r}$, by minimizing $\mathcal{L}_{\mathrm{total}}$ over samples drawn from a combined dataset $\mathcal{D}_{\mathrm{moe}}$ (constructed by sampling an equal number of examples from each the task in ${\{1,...N\}}$). Since each adapter’s cores are fixed,  interference  between experts is avoided, and the router learns to  to dispatch each $\mathbf{x}_b$ to its most suitable TT-LoRA expert, eliminating the need for manual adapter selection at inference.

\section{Experiments and Results}
\label{sec:experimentsandresults}
Our experimental pipeline consists of two stages: (i) training individual experts using parameter-efficient fine-tuning (PEFT) technique, and (ii) training the Mixture-of-Experts (MoE) router over the frozen experts.

The purpose of step (i) is to isolate the task knowledge from the base knowledge and also to ensure parametric and inference efficiency without compromising the performance. This aids in solving inter-task interference and knowledge forgetting issues at the same time it enhances the efficient multi-tasking capabilities. Furthermore, step (ii) helps to dynamically adapt the pre-trained expert during inference time and showcase the performance equivalent to individual expert's performance, actually solving the issue of  inter-task interference and knowledge forgetting during multi-tasking.

\subsection{Part I: Training Individual Experts}
\label{sec:trainingexperts}
Before training the experts for our multi-task MoE, we define clear criteria for selecting the most suitable PEFT adapters as expert modules. Specifically, our expert selection process is guided by three primary objectives:

\begin{enumerate} 
\item \textbf{Parameter Efficiency:} Selected experts should require significantly fewer trainable parameters. \item \textbf{Inference Efficiency:} Experts should enable rapid inference with minimal latency.
\item \textbf{Competitive Performance:} The performance of the experts should be on par with state-of-the-art methods. \end{enumerate}

\subsubsection{\textbf{Parameter Efficiency:}} As state‐of‐the‐art PEFT methods, we focus on the widely accepted LoRA~\cite{hu2021lora} and the more recent TT-LoRA~\cite{anjum2024tensor}. Anjum et al. report that TT-LoRA trains approximately 13 times fewer parameters than LoRA while achieving comparable performance. Consequently, to rigorously evaluate the parametric efficiency and performance of TT-LoRA relative to LoRA, we conducted an extensive comparative study.

\begin{table}[ht]
  \caption{Hyper-parameter search for LoRA and TT-LoRA}
  \label{tab:hyperparameters}
  \begin{tabular}{cccccl}
    \toprule
    Rank&Alpha&LR&TT-shape(Q/V)&Parameters&Acc.\\
    \midrule
    16&10&1e-3&-&1,703,936&0.87\\
    64&4&5e-4&-&6,815,744&0.88\\
    \textbf{16}&\textbf{8}&\textbf{5e-4}&-&\textbf{1,703,936}&\textbf{0.89}\\
    \toprule
    8&5&5e-3&[16,16,8,8,16,16]&98,304&0.873\\
    &&&[16,16,2,2,2,2,16,16]&&\\
    10&2&5e-4&[32, 8, 8, 8, 8, 32]&1,45,920&0.875\\
    &&&[64,32,8,64]&&\\
    \textbf{5}&\textbf{16}&\textbf{5e-3}&\textbf{[16,8,4,4,4,4,8,16]}&\textbf{33,920}&\textbf{0.875}\\
    &&&\textbf{[16,16,4,2,2,16,16]}&&\\
    \bottomrule
\end{tabular}
\end{table}

\textbf{TT-LoRA Training: }
We use \textbf{LlaMA-3.2-1B }as our base model. An individual TT-LoRA expert is trained by first loading a pre-trained base model and then augmenting the Query weights (\(2048\times2048\)) and Value weights (\(2048\times512\)) with $\Delta W$ for all 16 self-attention layers with untrained TT-cores (decomposed $\Delta W $ into Tensor-Train cores) weights. The TT-cores are initialized based on a chosen TT-shape comprising factors \([m_1,\dots,m_p,\,n_1,\dots,n_q]\) and TT-rank \([1, r, \dots, r, 1]\), effectively producing tensor cores of shape \([(m_i \text{ or } n_i)\times r_i \times (m_{i+1} \text{ or } n_{i+1})]\). During training, all base model weights are frozen except the TT-cores. Then the trained tensor cores and frozen classification head weights are preserved to be reused during router training.

\textbf{LoRA Training: } 
Similarly, using the same LlaMA-3.2-1B base model, individual LoRA experts are trained by wrapping the Query and Value weights (all 16 layers) with a custom LoRA class. In this wrapper, low-rank matrices (A and B) are initialized with a specified rank \(r\) and scaling factor \(\alpha\). All parameters, except these matrices, are frozen during training.

\textbf{Hyper-Parameter Search: } 
We conducted hyperparameter search via Ray Tune for both LoRA and TT-LoRA. For LoRA, the search tuned the learning rate, rank, and \(\alpha\); for TT-LoRA, it tuned the learning rate, TT-shape, rank, and \(\alpha\). Both were validated on the Microsoft Research Paraphrase Corpus (MRPC)~\cite{dolan2005automatically}. Table~\ref{tab:hyperparameters} shows the best three configurations for each method (LoRA in the upper half, TT-LoRA in the lower half), with bold parameters indicating the best trade-off between accuracy and trainable parameters. From Table ~\ref{tab:hyperparameters}, based on selected hyper-parameters, we can infer that TTLoRA method is extremely parametric efficient as compared to LoRA as it uses only \textbf{2\%} of the LoRA parameters count.

\begin{table}[ht]
  \caption{Inference Time Testing for TT-LoRA Adaptation}
  \label{tab:inferencetime}
  \begin{tabular}{cccl}
    \toprule
    Batch Size&Reconstruction& Input Contraction&Speed Up\\
    \midrule
    2 & 0.1125 sec & \textbf{0.0597 sec}&1.9x \\
    4 & 0.0689 sec & \textbf{0.0615 sec}&1.1x \\
    8 & 0.0667 sec & \textbf{0.0590 sec}&1.1x \\
    16 & 0.0789 sec & \textbf{0.0596 sec}&1.3x \\
    32 & 0.0713 sec & \textbf{0.0619 sec}&1.2x \\
    64 & 0.0783 sec & \textbf{0.0782 sec}&1.0x \\
    128 & 0.1032 sec & \textbf{0.0801 sec}&1.3x \\
    \bottomrule
\end{tabular}
\end{table}
\subsubsection{\textbf{Inference Efficiency:}}  
Along with the information about parametric efficiency of TT-LoRA, we conducted an experiment to observe the inference efficiency between the conventional reconstruction method and our proposed contraction method. The original TT-LoRA adaptation proposed by Anjum et al. \cite{anjum2024tensor} reconstructs the tensor-train (TT) cores into a full low-rank update matrix matching the shape of the original weight matrix (see Section~\ref{sec:background}). We contend that a more efficient alternative is to process the input via direct tensor contraction, as described in Section~\ref{subsec:tt-forward} and illustrated in Figure~\ref{fig:expert-training}. To validate this claim, we compared inference times for both methods across varying batch sizes, averaging the results over 10 runs (Table~\ref{tab:inferencetime}). The results clearly indicate an improvement in inference speed with our proposed contraction method. This ensures the our proposed input-tensor contraction based TT-LoRA method is inference efficient than existing TT-LoRA reconstruction method. Consequently, we adopt the contraction approach for training experts using TT-LoRA PEFT. 

\subsubsection{\textbf{Competitive Performance:}}
\label{sec:competitiveperformance}
Now that we have parameter efficient and inference efficient validation of the tensor contraction based TT-LoRA method, we perform extensive experiments to validate that it also delivers comparable performance with LoRA method.

\textbf{Datasets and Training:} With the best hyper-parameter configurations and efficient adaptation method, we trained both LoRA and TT-LoRA adapters on 17 distinct NLP based classification datasets, spanning diverse domains and task types: \textbf{Sentiment Analysis - }IMDB \cite{maas2011learning}, SST2 ~\cite{socher2013recursive}, \textbf{Commonsense Reasoning - } Hellaswag ~\cite{zellers2018swag}, Winogrande\_Large ~\cite{sakaguchi2021winogrande}, CosmosQA ~\cite{huang2019cosmos}, SocialIQA ~\cite{sap2019socialiqa}, CommonsenseQA ~\cite{talmor2018commonsenseqa}, \textbf{Natural Language Inference - }MultiNLI ~\cite{williams2017broad}, QNLI ~\cite{wang2018glue}, SciTail ~\cite{khot2018scitail}, SICK ~\cite{marelli-etal-2014-sick}, Recognizing Textual Entailment (RTE) ~\cite{wang2018glue}, Commitment Bank (CB) ~\cite{de2019commitmentbank}, \textbf{Sentence Relatedness - }Microsoft Research Paraphrase Corpus (MRPC) ~\cite{dolan2005automatically}, Quora Question Pairs (QQP) ~\cite{sharma2019natural}, \textbf{Question Answering - }BoolQ ~\cite{clark2019boolq}, \textbf{Grammatical Correctness - } CoLA ~\cite{warstadt2018neural}. In addition, the datasets in all the tables below are arranged in ascending order of the number of dataset samples present in the datasets. The tables categorize the datasets based on samples count <10,000, >10,000 but <100,000 and >100,000. The models were trained with patience value as 10 (if validation accuracy doesn't improve over 10 runs, the training terminates) and performance evaluation were carried out using validation set accuracy. We carried out the experiments using 4 80GB NVIDIA A100 GPUs for each training. The results for each dataset are presented in Table ~\ref{tab:loravsttlora}. 

\textbf{Performance vs. Parameter Count: } The results indicate that TT-LoRA achieves competitive performance, often matching or in some cases even exceeding LoRA, with only marginal differences in accuracy across most tasks. However, TT-LoRA demonstrates a substantial advantage in parameter efficiency—it uses roughly \textbf{2\% of the trainable parameters} (33,920) compared to LoRA's 1,703,936 parameters for the same adaptation layer. This significant reduction in parameter count underscores TT-LoRA’s potential as a scalable expert representation method, particularly in multi-task settings where the total number of experts can increase linearly with the number of tasks. Based on this analysis, we adopt the tensor contraction-based TT-LoRA PEFT method as the expert module in our subsequent MoE stage.

\subsection{Part II: Routing Evaluation \& MoE Training}
In the second stage, we integrate the individually trained 17 TT-LoRA experts into a unified MoE model, training only a lightweight router while keeping  the base model and all expert adapters frozen. 
This step ensures parameter efficient real-time adaptation of the pre-trained experts with the base model and allows us to evaluate how effectively our architecture mitigates inter-task interference and catastrophic forgetting. To achieve these goals, we focus on these three aspects:

\begin{enumerate}
\item \textbf{Parameter-Efficient and Accurate Routing:} Develop a lightweight router that selects the most appropriate expert with minimal additional parameters. \item \textbf{Retention of Individual Expert Performance:} Ensure that each expert maintains its pre-trained performance when incorporated into the MoE framework. 
\item \textbf{Effective Multi-Task Learning:} Validate the unified MoE model’s ability to perform across multiple tasks without degradation in performance.
\end{enumerate}

\begin{table}[ht]
  \caption{Performance Measures of LoRA vs. TT-LoRA}
  \label{tab:loravsttlora}
  \begin{tabular}{ccl}
    \toprule
    Datasets&LoRA&TT-LoRA\\
    \midrule
    RTE & \textbf{80.71} & \textbf{80.71} \\
    MRPC & \textbf{86.03} & 83.82 \\
    CB & 80.30 & \textbf{82.14} \\
    SICK & 90.90 & \textbf{91.12} \\
    CoLA & \textbf{84.96 }& 84.39 \\
    BoolQ & \textbf{77.81} & 68.58 \\
    CSQA & 21.16 & \textbf{22.55} \\
    \midrule
    Winogrande\_l & \textbf{65.06} & 54.73 \\
    SciTail & \textbf{96.86} & 95.46 \\
    IMDB & \textbf{96.16} & 95.43 \\
    CosmosQA & 72.19 &\textbf{74.33} \\
    SocialIQA & 71.93 & \textbf{73.36} \\
    Hellaswag & \textbf{84.96} & 84.00 \\
    SST2 & \textbf{95.98} & 95.76 \\
    \midrule
    QNLI & \textbf{92.42} & 91.36 \\
    QQP & \textbf{90.78} & 88.30 \\
    MNLI & \textbf{88.72} & 86.79 \\
    \midrule
    Average & \textbf{80.99} & 79.58 \\
    \bottomrule
\end{tabular}
\end{table}

\subsubsection{\textbf{Parameter-Efficient and Accurate Routing:}}
\label{sec:accuraterouting}
The architecture integrates pre-trained, parameter-efficient experts, so it is crucial to have a highly accurate router that correctly assigns the given input to its corresponding expert adapter. To fully leverage the performance of these experts, our design prioritizes both accuracy and minimal parameter overhead in the router. Consequently, we perform a systematic search to identify the router configuration that best meets these criteria.

\begin{table}[ht]
  \caption{Performance Measures of Different Router Types}
  \label{tab:routerselection}
  \begin{tabular}{ccccl}
    \# Tasks & Metrics & Single & Multi & Ours \\
    \midrule
    1,2 & Router Acc & 0.99 &\textbf{1.00}&\textbf{1.00} \\
    &Parameters &8,196&2,104,324&\textbf{8,194} \\
    \midrule
    1, 2, 3&Router Acc &0.75&0.66&\textbf{1.00}\\
    &Parameters&12,294&2,107,398&\textbf{12,291}\\
    \midrule
    1, 2, 3, 4&Router Acc&0.51&0.58&\textbf{1.00}\\
    &Parameters&16,392&2,110,472&\textbf{16,388}\\
    \midrule
    1, 2, 3, 4, 5&Router Acc&0.46&0.36&\textbf{0.99}\\
    &Parameters&20,490&2,113,546&\textbf{20,485}\\
    \midrule
    1, 2, 3, 4, 5, 6&Router Acc&0.45&0.28&\textbf{0.99}\\
    &Parameters&24,588&2,116,620&\textbf{24,582}\\
    \bottomrule
\end{tabular}
\end{table}

\begin{table*}[ht]
  \caption{Performance comparison between Adapter Fusion and TT-LoRA MoE on Single Datasets}
  \label{tab:singletaskmoecomparison}
  \begin{tabular}{c@{\hskip 12pt}cc@{\hskip 12pt}cc}
    \toprule
    \multirow{2}{*}{Dataset} & \multicolumn{2}{c}{\textbf{Single Training}} & \multicolumn{2}{c}{\textbf{Expert Integration}} \\
    \cmidrule(lr){2-3} \cmidrule(lr){4-5}
     & Pfeiffer Adapters & TTLoRA & AdapterFusion & TT-LoRA MoE \\
    \midrule
    RTE & \textbf{81.07} & 80.71 & 76.89 & \textbf{80.36} \\
    MRPC & \textbf{86.27} & 83.82 & \textbf{87.50} & 83.82 \\
    CB & \textbf{87.50} & 82.14 & 71.43 & \textbf{82.14} \\
    SICK & 90.60 & \textbf{91.12} & 90.80 & \textbf{91.06} \\
    CoLA & 83.31 & \textbf{84.39} & 83.32 & \textbf{84.37} \\
    BoolQ & \textbf{79.37} & 68.58 & \textbf{78.81} & 63.57 \\
    CSQA & 22.06 & \textbf{22.55} & 22.28 & \textbf{22.30} \\
    \midrule
    Winogrande\_l & \textbf{66.72} & 54.73 & \textbf{67.01} & 51.18 \\
    SciTail & \textbf{96.09} & 95.46 & \textbf{96.24} & 95.64 \\
    IMDB & \textbf{96.32} & 95.43 & \textbf{96.02} & 95.42 \\
    CosmosQA & 73.46 & \textbf{74.33} & 73.67 & \textbf{74.3} \\
    SocialIQA & 72.90 & \textbf{73.36} & 71.96 & \textbf{73.31} \\
    Hellaswag & \textbf{86.93} & 84.00 & \textbf{86.93} & 84.00 \\
    SST2 & 95.41 & \textbf{95.76} & 94.95 & \textbf{95.76} \\
    \midrule
    QNLI & \textbf{92.44} & 91.36 & \textbf{92.60} & 91.36 \\
    QQP & \textbf{91.34} & 88.30 & \textbf{91.19} & 88.22 \\
    MNLI & \textbf{89.32} & 86.79 & \textbf{87.81} & 86.79 \\
    \midrule
    Average & \textbf{81.83} & 79.58 & 75.16 & \textbf{79.04} \\
    \bottomrule
  \end{tabular}
\end{table*}

\textbf{Router Design Variants:}
We designed and evaluated three primary architectural choices for the router: (i) a single-layer linear router, (ii) a two-layer MLP router with ReLU non-linearity, and (iii) a single-layer trainable router that directly leverages the base model's hidden representations (ours). To assess \textbf{router accuracy} and compute \textbf{router loss}, during training we augmented our dataset with task labels corresponding to the expert indices (i.e., the index of the gate logits). More details about the integration of the router with the base model representations will be provided in the upcoming subsection.

The experiments were conducted incrementally by training the router—while keeping all other components frozen—with the number of experts gradually increased from 2 up to 6 (corresponding to the tasks: MNLI, CoLA, RTE, SST2, QNLI, and QQP). All router designs were evaluated on a \textbf{mixed dataset} constructed by sampling an equal number of instances from each task (based on the smallest dataset among the selected tasks). Our experiments consistently showed that the single-layer router leveraging the base model's final representation outperformed the other configurations. This indicates that the base model's representations inherently encode sufficient task-discriminative signals for effective expert selection. Detailed results are presented in Table~\ref{tab:routerselection}. In summary, the table demonstrates that our router is parameter efficient—with its parameter count increasing only by a constant factor as the number of experts grows—and highly accurate for up to 6 experts.

\textbf{Base Model Representations used for Routing the Expert:}
Figure~\ref{fig:moearchitecture} illustrates the complete TT-LoRA MoE architecture. First, the base model processes the input to produce contextual representations. Instead of forwarding these hidden states directly to the final projection layer, we pass them through a custom lightweight router that outputs gate logit scores. As described in Section~\ref{subsec:routing-base-model-leverage}, these scores are augmented with input-dependent trainable noise. The top-1 gate logit is then selected while the remaining logits are set to $- \infty$, and applying a softmax function produces a one-hot encoded gate vector.

The pre-trained TT-LoRA experts are organized into tensor stacks containing the TT-core tensors for the Q and K weights across multiple layers, while a separate stack maintains the classification or score layer weights of each expert. When the one-hot gate vector is obtained, it is applied to the expert and classification stacks, selecting the corresponding Q and V TT-cores for all 16 layers as well as the projection layer for the input task. These selected components are then combined with the base model, and the model subsequently performs a second forward pass for label prediction. This two-pass structure—first for routing and then for prediction—enables dynamic per-input expert selection and supports scalable multi-task adaptation. As depicted in Figure~\ref{fig:expert-training}, the selected expert’s tensor cores are contracted with the input to generate the final prediction.

\subsubsection{\textbf{Retention of Individual Expert Performance:}} From Section ~\ref{sec:competitiveperformance} and above subsection, we individually trained parametric efficient TT-LoRA experts with competitive performance along with highly accurate and low parameterized router. Thus, to test its effectiveness to retain the individual expert's performance, we perform a comparative analysis with the state-of-the-art AdapterFusion ~\cite{pfeiffer2020adapterfusion} method. In this method, multiple Pfeiffer ~\cite{pfeiffer2020adapterhub} adapters are trained separately in different NLP based classification tasks. Then, the saved adapters are loaded and fused together with the help of attention mechanism to introduce multi-task ability in the base model. During the training of the fusion architecture, all the weights are frozen except the fusion layer. 

\textbf{Comparison with AdapterFusion:}  
AdapterFusion ~\cite{pfeiffer2020adapterfusion} is the most directly comparable method, as it also targets modular reuse of pre-trained adapters. However, unlike our approach—which trains a lightweight router to dynamically select one expert—the AdapterFusion strategy learns a fusion layer that jointly attends to all adapters. This additional fusion layer increases both the parameter count and runtime cost. For a fair comparison, we first trained 17 different adapters on the datasets described in Section~\ref{sec:competitiveperformance} and saved the trained adapters. We then introduced a fusion layer to combine these 17 experts, training and evaluating the fusion on each dataset individually. For our TT-LoRA MoE, we also save the individual experts (as reported in Table~\ref{tab:loravsttlora}) and then introduce a trainable linear layer router fed with base model's representations. We then subsequently train the router layer, evaluating performance on each dataset one by one. 

\begin{table}[ht]
  \caption{Parameters and Accuracy Comparison}
  \label{tab:paravsaccuarcy}
  \begin{tabular}{cccl}
    \toprule
    Setting&Dataset Mix&Parameters&Avg. Acc.\\
    \midrule
    Single Adapters&Single&12,623,874&\textbf{81.83}\\
    Single TTLoRA&Single&\textbf{33,920}&79.58\\
    \midrule
    AdapterFusion&Single&205,592,578&75.16\\
    TT-LoRA MoE&Single&\textbf{69,649}&\textbf{79.04}\\
    \midrule
    AdapterFusion&Top 10 Mix&205,592,578&81.45\\
    TT-LoRA MoE&Top 10 Mix&\textbf{69,649}&\textbf{85.91}\\
    \bottomrule
\end{tabular}
\end{table}

\textbf{Comparison between Individual Experts:} Here, we discuss the performance comparison of individual separately trained PEFTs (experts). In the Table ~\ref{tab:singletaskmoecomparison}, under the Single Training heading, we can see that, TTLoRA's performance is competitive as compared to the Pfeiffer adapters, whereas in 6 out of 17 tasks our experts beat the adapter's performance. When compared the parameter count of Pfeiffer adapters i.e. 12,623,874 and ours i.e. 33,920, TT-LoRA utilizes approx. \textbf{0.3\%} of the adapters parameters yet it's average performance is nearly equal to the adapters with such high parameters count. 

\textbf{Comparison between Multi-task Experts Setting:} Once the individual experts are trained and saved separately, for Adapterfusion, we load all the 17 saved frozen adapters and initialize an untrained fusion layer. For our MoE model, we load all the 17 expert's tensor cores into a stack and then initialize an untrained single-layer-base-model-fed router. As mentioned above, we train the fusion and MoE with each dataset and note the performance in Table ~\ref{tab:singletaskmoecomparison}. Under the Expert Integration heading, we can see that our MoE model retains approximately equal performance as that of individual experts performance. It also beats the fusion's performance in 8 out of 17 tasks. We can also see that the average performance of our TT-LoRA MoE exceeds the fusion's performance by almost 4 points. When compared with the trainable parameters count in fusion and router layer which are: 205,592,578 for fusion and 69,649 for router, our architecture uses approx. \textbf{0.03\%} of the fusion layer's parameters, which is a huge saving in memory usage. 

From the experiments and results as discussed above, we can say that our proposed model retains the individual expert's performance in MoE setting in a very effective and efficient fashion. This also demonstrates that our proposed architecture also \textbf{solves} the issue of \textbf{inter-task interference} by retaining the separately trained individual experts' performance. 

\subsubsection{\textbf{Effective Multi-Task Learning:}}
Since our model performs effectively in single task setting when combined with 17 different experts routed with an accurate router, we also claim that it should perform as per the average of the combined experts with their corresponding datasets mix. So, we select the top 10 performing experts based on their accuracy from Table ~\ref{tab:singletaskmoecomparison}. The selected experts are: MRPC, SICK , CoLA, SciTail, IMDB, Hellaswag, SST2, QNLI, QQP, and MNLI. Thenafter, we load the dataset with the minimum count and slice other datasets based on it, mix the samples and shuffle them to train and evaluate the fusion and MoE setting. As presented in the last row of Table ~\ref{tab:paravsaccuarcy}, we can see that TT-LoRA MoE exceeds the performance of AdapterFusion by approx. 4 value with only using about \textbf{0.03\%} of the trainable parameters count of Fusion layer. This signifies that our proposed model can effectively learn and perform multi-tasking without loosing the performance of separately trained individual experts. 

From all the discussions above we see that, for an individual expert training-we only train the extra added PEFT weights, for MoE training-we only train the extra added router layer. In all these cases, we never change any weight from the base model. Hence, we also \textbf{solve knowledge forgetting} issue in multi-task learning environment with an efficient, effective and faster solution as TT-LoRA MoE. In addition, as mentioned under Router Design Variant at Section ~\ref{sec:accuraterouting}, the router can accurately classify 6 tasks and as shown above, our router implementation can accurately classify upto 17 experts with a constant growth in router parameter count. This clearly demonstrates that our TT-LoRA MoE design is parameter-efficient as well as posses scalable capacity. 

\section{Conclusion and Future Work}
We introduced \textbf{TT-LoRA MoE}, a modular and scalable Mixture-of-Experts architecture that addresses critical limitations in parameter-efficient fine-tuning and sparse expert integration. Our approach employs a two-stage pipeline: first, independently training highly compact and inference-efficient TT-LoRA experts; second, integrating these experts through a lightweight, noisy top-1 sparse router that leverages base model representations.

Through extensive experiments across \textbf{17 diverse NLP classification tasks}, we demonstrated that TT-LoRA experts achieve competitive performance while significantly reducing training overhead and memory footprint, using only \textbf{2\%} of the parameters compared to standard LoRA methods and only \textbf{0.3\%} of the adapters. Furthermore, our routing mechanism effectively retains individual expert performance and scales robustly in multi-task scenarios. TT-LoRA MoE outperformed AdapterFusion in 8 out of 17 tasks and achieved an average accuracy improvement of approximately \textbf{4 points} in both single-task and mixed-task evaluations, utilizing merely \textbf{0.03\%} of AdapterFusion’s trainable parameters.

By minimally fine-tuning task-specific TT-LoRA cores and a single-layer router for MoE training, our approach effectively preserves base model knowledge and avoids catastrophic forgetting. These attributes position TT-LoRA MoE as an attractive solution for real-world deployments requiring modular, reusable, and cost-efficient adaptation across multiple tasks. Overall, our contributions provide a highly parameter-efficient, scalable, and task-agnostic expert selection framework, laying strong foundations for future multi-task and continual learning systems.

While the current model trains experts and MoE components under homogeneous settings—uniform TT-shape, TT-rank, Alpha values, and task type (classification)—as part of future work, we plan to extend our architecture design toward heterogeneous scenarios involving variable TT-parameters and diverse task types (e.g., generation, regression, and question-answering). Additionally, we aim to explore token-level fusion of pre-trained experts to evaluate their inherent capabilities in unseen task conditions. Lastly, we intend to systematically investigate scalability limits beyond the current 17-expert configuration, extending toward hundreds of experts, to rigorously analyze performance trade-offs and identify potential architectural improvements.

 \section*{Acknowledgment}
  This research was funded by the LANL LDRD grant 20240777ER and the LANL Institutional Computing Program, supported by the U.S. Department of Energy National Nuclear Security Administration under Contract No 89233218CNA000001. The work is partially supported by the NSF grants 2230609 and 2416990 at Tennessee Tech. 

\bibliographystyle{ACM-Reference-Format}

\begin{thebibliography}{55}


\ifx \showCODEN    \undefined \def \showCODEN     #1{\unskip}     \fi
\ifx \showISBNx    \undefined \def \showISBNx     #1{\unskip}     \fi
\ifx \showISBNxiii \undefined \def \showISBNxiii  #1{\unskip}     \fi
\ifx \showISSN     \undefined \def \showISSN      #1{\unskip}     \fi
\ifx \showLCCN     \undefined \def \showLCCN      #1{\unskip}     \fi
\ifx \shownote     \undefined \def \shownote      #1{#1}          \fi
\ifx \showarticletitle \undefined \def \showarticletitle #1{#1}   \fi
\ifx \showURL      \undefined \def \showURL       {\relax}        \fi
\providecommand\bibfield[2]{#2}
\providecommand\bibinfo[2]{#2}
\providecommand\natexlab[1]{#1}
\providecommand\showeprint[2][]{arXiv:#2}

\bibitem[Anjum et~al\mbox{.}(2024)]%
        {anjum2024tensor}
\bibfield{author}{\bibinfo{person}{Afia Anjum}, \bibinfo{person}{Maksim~E. Eren}, \bibinfo{person}{Ismael Boureima}, \bibinfo{person}{Boian Alexandrov}, {and} \bibinfo{person}{Manish Bhattarai}.} \bibinfo{year}{2024}\natexlab{}.
\newblock \showarticletitle{Tensor Train Low-rank Approximation (TT-LoRA): Democratizing AI with Accelerated LLMs}. In \bibinfo{booktitle}{\emph{2024 International Conference on Machine Learning and Applications (ICMLA)}}. \bibinfo{pages}{583--590}.
\newblock
\href{https://doi.org/10.1109/ICMLA61862.2024.00085}{doi:\nolinkurl{10.1109/ICMLA61862.2024.00085}}


\bibitem[Brown et~al\mbox{.}(2020)]%
        {brown2020language}
\bibfield{author}{\bibinfo{person}{Tom Brown}, \bibinfo{person}{Benjamin Mann}, \bibinfo{person}{Nick Ryder}, \bibinfo{person}{Melanie Subbiah}, \bibinfo{person}{Jared~D Kaplan}, \bibinfo{person}{Prafulla Dhariwal}, \bibinfo{person}{Arvind Neelakantan}, \bibinfo{person}{Pranav Shyam}, \bibinfo{person}{Girish Sastry}, \bibinfo{person}{Amanda Askell}, {et~al\mbox{.}}} \bibinfo{year}{2020}\natexlab{}.
\newblock \showarticletitle{Language models are few-shot learners}.
\newblock \bibinfo{journal}{\emph{Advances in neural information processing systems}}  \bibinfo{volume}{33} (\bibinfo{year}{2020}), \bibinfo{pages}{1877--1901}.
\newblock


\bibitem[Clark et~al\mbox{.}(2019)]%
        {clark2019boolq}
\bibfield{author}{\bibinfo{person}{Christopher Clark}, \bibinfo{person}{Kenton Lee}, \bibinfo{person}{Ming-Wei Chang}, \bibinfo{person}{Tom Kwiatkowski}, \bibinfo{person}{Michael Collins}, {and} \bibinfo{person}{Kristina Toutanova}.} \bibinfo{year}{2019}\natexlab{}.
\newblock \showarticletitle{Boolq: Exploring the surprising difficulty of natural yes/no questions}.
\newblock \bibinfo{journal}{\emph{arXiv preprint arXiv:1905.10044}} (\bibinfo{year}{2019}).
\newblock


\bibitem[De~Marneffe et~al\mbox{.}(2019)]%
        {de2019commitmentbank}
\bibfield{author}{\bibinfo{person}{Marie-Catherine De~Marneffe}, \bibinfo{person}{Mandy Simons}, {and} \bibinfo{person}{Judith Tonhauser}.} \bibinfo{year}{2019}\natexlab{}.
\newblock \showarticletitle{The commitmentbank: Investigating projection in naturally occurring discourse}. In \bibinfo{booktitle}{\emph{proceedings of Sinn und Bedeutung}}, Vol.~\bibinfo{volume}{23}. \bibinfo{pages}{107--124}.
\newblock


\bibitem[Dettmers et~al\mbox{.}(2023)]%
        {dettmers2023qlora}
\bibfield{author}{\bibinfo{person}{Tim Dettmers}, \bibinfo{person}{Artidoro Pagnoni}, \bibinfo{person}{Ari Holtzman}, {and} \bibinfo{person}{Luke Zettlemoyer}.} \bibinfo{year}{2023}\natexlab{}.
\newblock \showarticletitle{Qlora: Efficient finetuning of quantized llms}.
\newblock \bibinfo{journal}{\emph{Advances in neural information processing systems}}  \bibinfo{volume}{36} (\bibinfo{year}{2023}), \bibinfo{pages}{10088--10115}.
\newblock


\bibitem[Dolan and Brockett(2005)]%
        {dolan2005automatically}
\bibfield{author}{\bibinfo{person}{Bill Dolan} {and} \bibinfo{person}{Chris Brockett}.} \bibinfo{year}{2005}\natexlab{}.
\newblock \showarticletitle{Automatically constructing a corpus of sentential paraphrases}. In \bibinfo{booktitle}{\emph{Third international workshop on paraphrasing (IWP2005)}}.
\newblock


\bibitem[Dou et~al\mbox{.}(2024)]%
        {dou2024loramoe}
\bibfield{author}{\bibinfo{person}{Shihan Dou}, \bibinfo{person}{Enyu Zhou}, \bibinfo{person}{Yan Liu}, \bibinfo{person}{Songyang Gao}, \bibinfo{person}{Wei Shen}, \bibinfo{person}{Limao Xiong}, \bibinfo{person}{Yuhao Zhou}, \bibinfo{person}{Xiao Wang}, \bibinfo{person}{Zhiheng Xi}, \bibinfo{person}{Xiaoran Fan}, {et~al\mbox{.}}} \bibinfo{year}{2024}\natexlab{}.
\newblock \showarticletitle{LoRAMoE: Alleviating world knowledge forgetting in large language models via MoE-style plugin}. In \bibinfo{booktitle}{\emph{Proceedings of the 62nd Annual Meeting of the Association for Computational Linguistics (Volume 1: Long Papers)}}. \bibinfo{pages}{1932--1945}.
\newblock


\bibitem[Du et~al\mbox{.}(2022)]%
        {du2022glam}
\bibfield{author}{\bibinfo{person}{Nan Du}, \bibinfo{person}{Yanping Huang}, \bibinfo{person}{Andrew~M Dai}, \bibinfo{person}{Simon Tong}, \bibinfo{person}{Dmitry Lepikhin}, \bibinfo{person}{Yuanzhong Xu}, \bibinfo{person}{Maxim Krikun}, \bibinfo{person}{Yanqi Zhou}, \bibinfo{person}{Adams~Wei Yu}, \bibinfo{person}{Orhan Firat}, {et~al\mbox{.}}} \bibinfo{year}{2022}\natexlab{}.
\newblock \showarticletitle{Glam: Efficient scaling of language models with mixture-of-experts}. In \bibinfo{booktitle}{\emph{International conference on machine learning}}. PMLR, \bibinfo{pages}{5547--5569}.
\newblock


\bibitem[Fedus et~al\mbox{.}(2022)]%
        {fedus2022switch}
\bibfield{author}{\bibinfo{person}{William Fedus}, \bibinfo{person}{Barret Zoph}, {and} \bibinfo{person}{Noam Shazeer}.} \bibinfo{year}{2022}\natexlab{}.
\newblock \showarticletitle{Switch transformers: Scaling to trillion parameter models with simple and efficient sparsity}.
\newblock \bibinfo{journal}{\emph{Journal of Machine Learning Research}} \bibinfo{volume}{23}, \bibinfo{number}{120} (\bibinfo{year}{2022}), \bibinfo{pages}{1--39}.
\newblock


\bibitem[French(1999)]%
        {french1999catastrophic}
\bibfield{author}{\bibinfo{person}{Robert~M French}.} \bibinfo{year}{1999}\natexlab{}.
\newblock \showarticletitle{Catastrophic forgetting in connectionist networks}.
\newblock \bibinfo{journal}{\emph{Trends in cognitive sciences}} \bibinfo{volume}{3}, \bibinfo{number}{4} (\bibinfo{year}{1999}), \bibinfo{pages}{128--135}.
\newblock


\bibitem[Gao et~al\mbox{.}(2024)]%
        {gao2024higher}
\bibfield{author}{\bibinfo{person}{Chongyang Gao}, \bibinfo{person}{Kezhen Chen}, \bibinfo{person}{Jinmeng Rao}, \bibinfo{person}{Baochen Sun}, \bibinfo{person}{Ruibo Liu}, \bibinfo{person}{Daiyi Peng}, \bibinfo{person}{Yawen Zhang}, \bibinfo{person}{Xiaoyuan Guo}, \bibinfo{person}{Jie Yang}, {and} \bibinfo{person}{VS Subrahmanian}.} \bibinfo{year}{2024}\natexlab{}.
\newblock \showarticletitle{Higher layers need more lora experts}.
\newblock \bibinfo{journal}{\emph{arXiv preprint arXiv:2402.08562}} (\bibinfo{year}{2024}).
\newblock


\bibitem[Gou et~al\mbox{.}(2023)]%
        {gou2023mixture}
\bibfield{author}{\bibinfo{person}{Yunhao Gou}, \bibinfo{person}{Zhili Liu}, \bibinfo{person}{Kai Chen}, \bibinfo{person}{Lanqing Hong}, \bibinfo{person}{Hang Xu}, \bibinfo{person}{Aoxue Li}, \bibinfo{person}{Dit-Yan Yeung}, \bibinfo{person}{James~T Kwok}, {and} \bibinfo{person}{Yu Zhang}.} \bibinfo{year}{2023}\natexlab{}.
\newblock \showarticletitle{Mixture of cluster-conditional lora experts for vision-language instruction tuning}.
\newblock \bibinfo{journal}{\emph{arXiv preprint arXiv:2312.12379}} (\bibinfo{year}{2023}).
\newblock


\bibitem[Houlsby et~al\mbox{.}(2019)]%
        {houlsby2019parameter}
\bibfield{author}{\bibinfo{person}{Neil Houlsby}, \bibinfo{person}{Andrei Giurgiu}, \bibinfo{person}{Stanislaw Jastrzebski}, \bibinfo{person}{Bruna Morrone}, \bibinfo{person}{Quentin De~Laroussilhe}, \bibinfo{person}{Andrea Gesmundo}, \bibinfo{person}{Mona Attariyan}, {and} \bibinfo{person}{Sylvain Gelly}.} \bibinfo{year}{2019}\natexlab{}.
\newblock \showarticletitle{Parameter-efficient transfer learning for NLP}. In \bibinfo{booktitle}{\emph{International conference on machine learning}}. PMLR, \bibinfo{pages}{2790--2799}.
\newblock


\bibitem[Hu et~al\mbox{.}(2022)]%
        {hu2021lora}
\bibfield{author}{\bibinfo{person}{Edward~J. Hu}, \bibinfo{person}{Yelong Shen}, \bibinfo{person}{Phillip Wallis}, \bibinfo{person}{Zeyuan Allen-Zhu}, \bibinfo{person}{Yuanzhong Xu}, \bibinfo{person}{Lu Wang}, {and} \bibinfo{person}{Weizhu Chen}.} \bibinfo{year}{2022}\natexlab{}.
\newblock \showarticletitle{LoRA: Low-Rank Adaptation of Large Language Models}. In \bibinfo{booktitle}{\emph{International Conference on Learning Representations (ICLR)}}.
\newblock
\urldef\tempurl%
\url{https://arxiv.org/abs/2106.09685}
\showURL{%
\tempurl}


\bibitem[Huang et~al\mbox{.}(2023)]%
        {huang2023lorahub}
\bibfield{author}{\bibinfo{person}{Chengsong Huang}, \bibinfo{person}{Qian Liu}, \bibinfo{person}{Bill~Yuchen Lin}, \bibinfo{person}{Tianyu Pang}, \bibinfo{person}{Chao Du}, {and} \bibinfo{person}{Min Lin}.} \bibinfo{year}{2023}\natexlab{}.
\newblock \showarticletitle{Lorahub: Efficient cross-task generalization via dynamic lora composition}.
\newblock \bibinfo{journal}{\emph{arXiv preprint arXiv:2307.13269}} (\bibinfo{year}{2023}).
\newblock


\bibitem[Huang et~al\mbox{.}(2019)]%
        {huang2019cosmos}
\bibfield{author}{\bibinfo{person}{Lifu Huang}, \bibinfo{person}{Ronan~Le Bras}, \bibinfo{person}{Chandra Bhagavatula}, {and} \bibinfo{person}{Yejin Choi}.} \bibinfo{year}{2019}\natexlab{}.
\newblock \showarticletitle{Cosmos QA: Machine reading comprehension with contextual commonsense reasoning}.
\newblock \bibinfo{journal}{\emph{arXiv preprint arXiv:1909.00277}} (\bibinfo{year}{2019}).
\newblock


\bibitem[Jacobs et~al\mbox{.}(1991)]%
        {jacobs1991adaptive}
\bibfield{author}{\bibinfo{person}{Robert~A Jacobs}, \bibinfo{person}{Michael~I Jordan}, \bibinfo{person}{Steven~J Nowlan}, {and} \bibinfo{person}{Geoffrey~E Hinton}.} \bibinfo{year}{1991}\natexlab{}.
\newblock \showarticletitle{Adaptive mixtures of local experts}.
\newblock \bibinfo{journal}{\emph{Neural computation}} \bibinfo{volume}{3}, \bibinfo{number}{1} (\bibinfo{year}{1991}), \bibinfo{pages}{79--87}.
\newblock


\bibitem[Jordan and Jacobs(1994)]%
        {jordan1994hierarchical}
\bibfield{author}{\bibinfo{person}{Michael~I Jordan} {and} \bibinfo{person}{Robert~A Jacobs}.} \bibinfo{year}{1994}\natexlab{}.
\newblock \showarticletitle{Hierarchical mixtures of experts and the EM algorithm}.
\newblock \bibinfo{journal}{\emph{Neural computation}} \bibinfo{volume}{6}, \bibinfo{number}{2} (\bibinfo{year}{1994}), \bibinfo{pages}{181--214}.
\newblock


\bibitem[Kaplan et~al\mbox{.}(2020)]%
        {kaplan2020scaling}
\bibfield{author}{\bibinfo{person}{Jared Kaplan}, \bibinfo{person}{Sam McCandlish}, \bibinfo{person}{Tom Henighan}, \bibinfo{person}{Tom~B Brown}, \bibinfo{person}{Benjamin Chess}, \bibinfo{person}{Rewon Child}, \bibinfo{person}{Scott Gray}, \bibinfo{person}{Alec Radford}, \bibinfo{person}{Jeffrey Wu}, {and} \bibinfo{person}{Dario Amodei}.} \bibinfo{year}{2020}\natexlab{}.
\newblock \showarticletitle{Scaling laws for neural language models}.
\newblock \bibinfo{journal}{\emph{arXiv preprint arXiv:2001.08361}} (\bibinfo{year}{2020}).
\newblock


\bibitem[Khot et~al\mbox{.}(2018)]%
        {khot2018scitail}
\bibfield{author}{\bibinfo{person}{Tushar Khot}, \bibinfo{person}{Ashish Sabharwal}, {and} \bibinfo{person}{Peter Clark}.} \bibinfo{year}{2018}\natexlab{}.
\newblock \showarticletitle{Scitail: A textual entailment dataset from science question answering}. In \bibinfo{booktitle}{\emph{Proceedings of the AAAI conference on artificial intelligence}}, Vol.~\bibinfo{volume}{32}.
\newblock


\bibitem[Lester et~al\mbox{.}(2021)]%
        {lester2021power}
\bibfield{author}{\bibinfo{person}{Brian Lester}, \bibinfo{person}{Rami Al-Rfou}, {and} \bibinfo{person}{Noah Constant}.} \bibinfo{year}{2021}\natexlab{}.
\newblock \showarticletitle{The power of scale for parameter-efficient prompt tuning}.
\newblock \bibinfo{journal}{\emph{arXiv preprint arXiv:2104.08691}} (\bibinfo{year}{2021}).
\newblock


\bibitem[Li et~al\mbox{.}(2024)]%
        {li2024mixlora}
\bibfield{author}{\bibinfo{person}{Dengchun Li}, \bibinfo{person}{Yingzi Ma}, \bibinfo{person}{Naizheng Wang}, \bibinfo{person}{Zhengmao Ye}, \bibinfo{person}{Zhiyuan Cheng}, \bibinfo{person}{Yinghao Tang}, \bibinfo{person}{Yan Zhang}, \bibinfo{person}{Lei Duan}, \bibinfo{person}{Jie Zuo}, \bibinfo{person}{Cal Yang}, {et~al\mbox{.}}} \bibinfo{year}{2024}\natexlab{}.
\newblock \showarticletitle{Mixlora: Enhancing large language models fine-tuning with lora-based mixture of experts}.
\newblock \bibinfo{journal}{\emph{arXiv preprint arXiv:2404.15159}} (\bibinfo{year}{2024}).
\newblock


\bibitem[Li and Liang(2021)]%
        {li2021prefix}
\bibfield{author}{\bibinfo{person}{Xiang~Lisa Li} {and} \bibinfo{person}{Percy Liang}.} \bibinfo{year}{2021}\natexlab{}.
\newblock \showarticletitle{Prefix-tuning: Optimizing continuous prompts for generation}.
\newblock \bibinfo{journal}{\emph{arXiv preprint arXiv:2101.00190}} (\bibinfo{year}{2021}).
\newblock


\bibitem[Liu et~al\mbox{.}(2023)]%
        {liu2023moelora}
\bibfield{author}{\bibinfo{person}{Qidong Liu}, \bibinfo{person}{Xian Wu}, \bibinfo{person}{Xiangyu Zhao}, \bibinfo{person}{Yuanshao Zhu}, \bibinfo{person}{Derong Xu}, \bibinfo{person}{Feng Tian}, {and} \bibinfo{person}{Yefeng Zheng}.} \bibinfo{year}{2023}\natexlab{}.
\newblock \showarticletitle{Moelora: An moe-based parameter efficient fine-tuning method for multi-task medical applications}.
\newblock \bibinfo{journal}{\emph{CoRR}} (\bibinfo{year}{2023}).
\newblock


\bibitem[Liu et~al\mbox{.}(2024)]%
        {liu2024perft}
\bibfield{author}{\bibinfo{person}{Yilun Liu}, \bibinfo{person}{Yunpu Ma}, \bibinfo{person}{Shuo Chen}, \bibinfo{person}{Zifeng Ding}, \bibinfo{person}{Bailan He}, \bibinfo{person}{Zhen Han}, {and} \bibinfo{person}{Volker Tresp}.} \bibinfo{year}{2024}\natexlab{}.
\newblock \showarticletitle{PERFT: Parameter-Efficient Routed Fine-Tuning for Mixture-of-Expert Model}.
\newblock \bibinfo{journal}{\emph{arXiv preprint arXiv:2411.08212}} (\bibinfo{year}{2024}).
\newblock


\bibitem[Liu and Luo(2024)]%
        {liu2024adamole}
\bibfield{author}{\bibinfo{person}{Zefang Liu} {and} \bibinfo{person}{Jiahua Luo}.} \bibinfo{year}{2024}\natexlab{}.
\newblock \showarticletitle{Adamole: Fine-tuning large language models with adaptive mixture of low-rank adaptation experts}.
\newblock \bibinfo{journal}{\emph{arXiv preprint arXiv:2405.00361}} (\bibinfo{year}{2024}).
\newblock


\bibitem[Maas et~al\mbox{.}(2011)]%
        {maas2011learning}
\bibfield{author}{\bibinfo{person}{Andrew Maas}, \bibinfo{person}{Raymond~E Daly}, \bibinfo{person}{Peter~T Pham}, \bibinfo{person}{Dan Huang}, \bibinfo{person}{Andrew~Y Ng}, {and} \bibinfo{person}{Christopher Potts}.} \bibinfo{year}{2011}\natexlab{}.
\newblock \showarticletitle{Learning word vectors for sentiment analysis}. In \bibinfo{booktitle}{\emph{Proceedings of the 49th annual meeting of the association for computational linguistics: Human language technologies}}. \bibinfo{pages}{142--150}.
\newblock


\bibitem[Marelli et~al\mbox{.}(2014)]%
        {marelli-etal-2014-sick}
\bibfield{author}{\bibinfo{person}{Marco Marelli}, \bibinfo{person}{Stefano Menini}, \bibinfo{person}{Marco Baroni}, \bibinfo{person}{Luisa Bentivogli}, \bibinfo{person}{Raffaella Bernardi}, {and} \bibinfo{person}{Roberto Zamparelli}.} \bibinfo{year}{2014}\natexlab{}.
\newblock \showarticletitle{A {SICK} cure for the evaluation of compositional distributional semantic models}. In \bibinfo{booktitle}{\emph{Proceedings of the Ninth International Conference on Language Resources and Evaluation ({LREC}`14)}}, \bibfield{editor}{\bibinfo{person}{Nicoletta Calzolari}, \bibinfo{person}{Khalid Choukri}, \bibinfo{person}{Thierry Declerck}, \bibinfo{person}{Hrafn Loftsson}, \bibinfo{person}{Bente Maegaard}, \bibinfo{person}{Joseph Mariani}, \bibinfo{person}{Asuncion Moreno}, \bibinfo{person}{Jan Odijk}, {and} \bibinfo{person}{Stelios Piperidis}} (Eds.). \bibinfo{publisher}{European Language Resources Association (ELRA)}, \bibinfo{address}{Reykjavik, Iceland}, \bibinfo{pages}{216--223}.
\newblock
\urldef\tempurl%
\url{https://aclanthology.org/L14-1314/}
\showURL{%
\tempurl}


\bibitem[McCloskey and Cohen(1989)]%
        {mccloskey1989catastrophic}
\bibfield{author}{\bibinfo{person}{Michael McCloskey} {and} \bibinfo{person}{Neal~J Cohen}.} \bibinfo{year}{1989}\natexlab{}.
\newblock \showarticletitle{Catastrophic interference in connectionist networks: The sequential learning problem}.
\newblock In \bibinfo{booktitle}{\emph{Psychology of learning and motivation}}. Vol.~\bibinfo{volume}{24}. \bibinfo{publisher}{Elsevier}, \bibinfo{pages}{109--165}.
\newblock


\bibitem[Misra et~al\mbox{.}(2016)]%
        {misra2016cross}
\bibfield{author}{\bibinfo{person}{Ishan Misra}, \bibinfo{person}{Abhinav Shrivastava}, \bibinfo{person}{Abhinav Gupta}, {and} \bibinfo{person}{Martial Hebert}.} \bibinfo{year}{2016}\natexlab{}.
\newblock \showarticletitle{Cross-stitch networks for multi-task learning}. In \bibinfo{booktitle}{\emph{Proceedings of the IEEE conference on computer vision and pattern recognition}}. \bibinfo{pages}{3994--4003}.
\newblock


\bibitem[Pfeiffer et~al\mbox{.}(2020a)]%
        {pfeiffer2020adapterfusion}
\bibfield{author}{\bibinfo{person}{Jonas Pfeiffer}, \bibinfo{person}{Aishwarya Kamath}, \bibinfo{person}{Andreas R{\"u}ckl{\'e}}, \bibinfo{person}{Kyunghyun Cho}, {and} \bibinfo{person}{Iryna Gurevych}.} \bibinfo{year}{2020}\natexlab{a}.
\newblock \showarticletitle{Adapterfusion: Non-destructive task composition for transfer learning}.
\newblock \bibinfo{journal}{\emph{arXiv preprint arXiv:2005.00247}} (\bibinfo{year}{2020}).
\newblock


\bibitem[Pfeiffer et~al\mbox{.}(2020b)]%
        {pfeiffer2020adapterhub}
\bibfield{author}{\bibinfo{person}{Jonas Pfeiffer}, \bibinfo{person}{Andreas R{\"u}ckl{\'e}}, \bibinfo{person}{Clifton Poth}, \bibinfo{person}{Aishwarya Kamath}, \bibinfo{person}{Ivan Vuli{\'c}}, \bibinfo{person}{Sebastian Ruder}, \bibinfo{person}{Kyunghyun Cho}, {and} \bibinfo{person}{Iryna Gurevych}.} \bibinfo{year}{2020}\natexlab{b}.
\newblock \showarticletitle{Adapterhub: A framework for adapting transformers}.
\newblock \bibinfo{journal}{\emph{arXiv preprint arXiv:2007.07779}} (\bibinfo{year}{2020}).
\newblock


\bibitem[Phang et~al\mbox{.}(2018)]%
        {phang2018sentence}
\bibfield{author}{\bibinfo{person}{Jason Phang}, \bibinfo{person}{Thibault F{\'e}vry}, {and} \bibinfo{person}{Samuel~R Bowman}.} \bibinfo{year}{2018}\natexlab{}.
\newblock \showarticletitle{Sentence encoders on stilts: Supplementary training on intermediate labeled-data tasks}.
\newblock \bibinfo{journal}{\emph{arXiv preprint arXiv:1811.01088}} (\bibinfo{year}{2018}).
\newblock


\bibitem[Pruksachatkun et~al\mbox{.}(2020)]%
        {pruksachatkun2020intermediate}
\bibfield{author}{\bibinfo{person}{Yada Pruksachatkun}, \bibinfo{person}{Jason Phang}, \bibinfo{person}{Haokun Liu}, \bibinfo{person}{Phu~Mon Htut}, \bibinfo{person}{Xiaoyi Zhang}, \bibinfo{person}{Richard~Yuanzhe Pang}, \bibinfo{person}{Clara Vania}, \bibinfo{person}{Katharina Kann}, {and} \bibinfo{person}{Samuel~R Bowman}.} \bibinfo{year}{2020}\natexlab{}.
\newblock \showarticletitle{Intermediate-task transfer learning with pretrained models for natural language understanding: When and why does it work?}
\newblock \bibinfo{journal}{\emph{arXiv preprint arXiv:2005.00628}} (\bibinfo{year}{2020}).
\newblock


\bibitem[Radford et~al\mbox{.}(2018)]%
        {radford2018improving}
\bibfield{author}{\bibinfo{person}{Alec Radford}, \bibinfo{person}{Karthik Narasimhan}, \bibinfo{person}{Tim Salimans}, \bibinfo{person}{Ilya Sutskever}, {et~al\mbox{.}}} \bibinfo{year}{2018}\natexlab{}.
\newblock \showarticletitle{Improving language understanding by generative pre-training}.
\newblock  (\bibinfo{year}{2018}).
\newblock


\bibitem[Radford et~al\mbox{.}(2019)]%
        {radford2019language}
\bibfield{author}{\bibinfo{person}{Alec Radford}, \bibinfo{person}{Jeffrey Wu}, \bibinfo{person}{Rewon Child}, \bibinfo{person}{David Luan}, \bibinfo{person}{Dario Amodei}, \bibinfo{person}{Ilya Sutskever}, {et~al\mbox{.}}} \bibinfo{year}{2019}\natexlab{}.
\newblock \showarticletitle{Language models are unsupervised multitask learners}.
\newblock \bibinfo{journal}{\emph{OpenAI blog}} \bibinfo{volume}{1}, \bibinfo{number}{8} (\bibinfo{year}{2019}), \bibinfo{pages}{9}.
\newblock


\bibitem[Rosenfeld and Tsotsos(2018)]%
        {rosenfeld2018incremental}
\bibfield{author}{\bibinfo{person}{Amir Rosenfeld} {and} \bibinfo{person}{John~K Tsotsos}.} \bibinfo{year}{2018}\natexlab{}.
\newblock \showarticletitle{Incremental learning through deep adaptation}.
\newblock \bibinfo{journal}{\emph{IEEE transactions on pattern analysis and machine intelligence}} \bibinfo{volume}{42}, \bibinfo{number}{3} (\bibinfo{year}{2018}), \bibinfo{pages}{651--663}.
\newblock


\bibitem[Sakaguchi et~al\mbox{.}(2021)]%
        {sakaguchi2021winogrande}
\bibfield{author}{\bibinfo{person}{Keisuke Sakaguchi}, \bibinfo{person}{Ronan~Le Bras}, \bibinfo{person}{Chandra Bhagavatula}, {and} \bibinfo{person}{Yejin Choi}.} \bibinfo{year}{2021}\natexlab{}.
\newblock \showarticletitle{Winogrande: An adversarial winograd schema challenge at scale}.
\newblock \bibinfo{journal}{\emph{Commun. ACM}} \bibinfo{volume}{64}, \bibinfo{number}{9} (\bibinfo{year}{2021}), \bibinfo{pages}{99--106}.
\newblock


\bibitem[Sap et~al\mbox{.}(2019)]%
        {sap2019socialiqa}
\bibfield{author}{\bibinfo{person}{Maarten Sap}, \bibinfo{person}{Hannah Rashkin}, \bibinfo{person}{Derek Chen}, \bibinfo{person}{Ronan LeBras}, {and} \bibinfo{person}{Yejin Choi}.} \bibinfo{year}{2019}\natexlab{}.
\newblock \showarticletitle{Socialiqa: Commonsense reasoning about social interactions}.
\newblock \bibinfo{journal}{\emph{arXiv preprint arXiv:1904.09728}} (\bibinfo{year}{2019}).
\newblock


\bibitem[Sharma et~al\mbox{.}(2019)]%
        {sharma2019natural}
\bibfield{author}{\bibinfo{person}{Lakshay Sharma}, \bibinfo{person}{Laura Graesser}, \bibinfo{person}{Nikita Nangia}, {and} \bibinfo{person}{Utku Evci}.} \bibinfo{year}{2019}\natexlab{}.
\newblock \showarticletitle{Natural language understanding with the quora question pairs dataset}.
\newblock \bibinfo{journal}{\emph{arXiv preprint arXiv:1907.01041}} (\bibinfo{year}{2019}).
\newblock


\bibitem[Shazeer et~al\mbox{.}(2017)]%
        {shazeer2017outrageously}
\bibfield{author}{\bibinfo{person}{Noam Shazeer}, \bibinfo{person}{Azalia Mirhoseini}, \bibinfo{person}{Krzysztof Maziarz}, \bibinfo{person}{Andy Davis}, \bibinfo{person}{Quoc Le}, \bibinfo{person}{Geoffrey Hinton}, {and} \bibinfo{person}{Jeff Dean}.} \bibinfo{year}{2017}\natexlab{}.
\newblock \showarticletitle{Outrageously Large Neural Networks: The Sparsely-Gated Mixture-of-Experts Layer}. In \bibinfo{booktitle}{\emph{International Conference on Learning Representations (ICLR)}}.
\newblock
\urldef\tempurl%
\url{https://arxiv.org/abs/1701.06538}
\showURL{%
\tempurl}


\bibitem[Socher et~al\mbox{.}(2013)]%
        {socher2013recursive}
\bibfield{author}{\bibinfo{person}{Richard Socher}, \bibinfo{person}{Alex Perelygin}, \bibinfo{person}{Jean Wu}, \bibinfo{person}{Jason Chuang}, \bibinfo{person}{Christopher~D Manning}, \bibinfo{person}{Andrew~Y Ng}, {and} \bibinfo{person}{Christopher Potts}.} \bibinfo{year}{2013}\natexlab{}.
\newblock \showarticletitle{Recursive deep models for semantic compositionality over a sentiment treebank}. In \bibinfo{booktitle}{\emph{Proceedings of the 2013 conference on empirical methods in natural language processing}}. \bibinfo{pages}{1631--1642}.
\newblock


\bibitem[Talmor et~al\mbox{.}(2018)]%
        {talmor2018commonsenseqa}
\bibfield{author}{\bibinfo{person}{Alon Talmor}, \bibinfo{person}{Jonathan Herzig}, \bibinfo{person}{Nicholas Lourie}, {and} \bibinfo{person}{Jonathan Berant}.} \bibinfo{year}{2018}\natexlab{}.
\newblock \showarticletitle{Commonsenseqa: A question answering challenge targeting commonsense knowledge}.
\newblock \bibinfo{journal}{\emph{arXiv preprint arXiv:1811.00937}} (\bibinfo{year}{2018}).
\newblock


\bibitem[Vaswani et~al\mbox{.}(2017)]%
        {vaswani2017attention}
\bibfield{author}{\bibinfo{person}{Ashish Vaswani}, \bibinfo{person}{Noam Shazeer}, \bibinfo{person}{Niki Parmar}, \bibinfo{person}{Jakob Uszkoreit}, \bibinfo{person}{Llion Jones}, \bibinfo{person}{Aidan~N Gomez}, \bibinfo{person}{{\L}ukasz Kaiser}, {and} \bibinfo{person}{Illia Polosukhin}.} \bibinfo{year}{2017}\natexlab{}.
\newblock \showarticletitle{Attention is all you need}.
\newblock \bibinfo{journal}{\emph{Advances in neural information processing systems}}  \bibinfo{volume}{30} (\bibinfo{year}{2017}).
\newblock


\bibitem[Wang et~al\mbox{.}(2018)]%
        {wang2018glue}
\bibfield{author}{\bibinfo{person}{Alex Wang}, \bibinfo{person}{Amanpreet Singh}, \bibinfo{person}{Julian Michael}, \bibinfo{person}{Felix Hill}, \bibinfo{person}{Omer Levy}, {and} \bibinfo{person}{Samuel~R Bowman}.} \bibinfo{year}{2018}\natexlab{}.
\newblock \showarticletitle{GLUE: A multi-task benchmark and analysis platform for natural language understanding}.
\newblock \bibinfo{journal}{\emph{arXiv preprint arXiv:1804.07461}} (\bibinfo{year}{2018}).
\newblock


\bibitem[Wang et~al\mbox{.}(2024)]%
        {wang2024malora}
\bibfield{author}{\bibinfo{person}{Xujia Wang}, \bibinfo{person}{Haiyan Zhao}, \bibinfo{person}{Shuo Wang}, \bibinfo{person}{Hanqing Wang}, {and} \bibinfo{person}{Zhiyuan Liu}.} \bibinfo{year}{2024}\natexlab{}.
\newblock \showarticletitle{MALoRA: Mixture of Asymmetric Low-Rank Adaptation for Enhanced Multi-Task Learning}.
\newblock \bibinfo{journal}{\emph{arXiv preprint arXiv:2410.22782}} (\bibinfo{year}{2024}).
\newblock


\bibitem[Wang et~al\mbox{.}(2022)]%
        {wang2022adamix}
\bibfield{author}{\bibinfo{person}{Yaqing Wang}, \bibinfo{person}{Subhabrata Mukherjee}, \bibinfo{person}{Xiaodong Liu}, \bibinfo{person}{Jing Gao}, \bibinfo{person}{Ahmed~Hassan Awadallah}, {and} \bibinfo{person}{Jianfeng Gao}.} \bibinfo{year}{2022}\natexlab{}.
\newblock \showarticletitle{Adamix: Mixture-of-adapter for parameter-efficient tuning of large language models}.
\newblock \bibinfo{journal}{\emph{arXiv preprint arXiv:2205.12410}} \bibinfo{volume}{1}, \bibinfo{number}{2} (\bibinfo{year}{2022}), \bibinfo{pages}{4}.
\newblock


\bibitem[Warstadt et~al\mbox{.}(2018)]%
        {warstadt2018neural}
\bibfield{author}{\bibinfo{person}{Alex Warstadt}, \bibinfo{person}{Amanpreet Singh}, {and} \bibinfo{person}{Samuel~R Bowman}.} \bibinfo{year}{2018}\natexlab{}.
\newblock \showarticletitle{Neural Network Acceptability Judgments}.
\newblock \bibinfo{journal}{\emph{arXiv preprint arXiv:1805.12471}} (\bibinfo{year}{2018}).
\newblock


\bibitem[Williams et~al\mbox{.}(2017)]%
        {williams2017broad}
\bibfield{author}{\bibinfo{person}{Adina Williams}, \bibinfo{person}{Nikita Nangia}, {and} \bibinfo{person}{Samuel~R Bowman}.} \bibinfo{year}{2017}\natexlab{}.
\newblock \showarticletitle{A broad-coverage challenge corpus for sentence understanding through inference}.
\newblock \bibinfo{journal}{\emph{arXiv preprint arXiv:1704.05426}} (\bibinfo{year}{2017}).
\newblock


\bibitem[Wu et~al\mbox{.}(2024)]%
        {wu2024mixture}
\bibfield{author}{\bibinfo{person}{Xun Wu}, \bibinfo{person}{Shaohan Huang}, {and} \bibinfo{person}{Furu Wei}.} \bibinfo{year}{2024}\natexlab{}.
\newblock \showarticletitle{Mixture of lora experts}.
\newblock \bibinfo{journal}{\emph{arXiv preprint arXiv:2404.13628}} (\bibinfo{year}{2024}).
\newblock


\bibitem[Yang et~al\mbox{.}(2024)]%
        {yang2024moral}
\bibfield{author}{\bibinfo{person}{Shu Yang}, \bibinfo{person}{Muhammad~Asif Ali}, \bibinfo{person}{Cheng-Long Wang}, \bibinfo{person}{Lijie Hu}, {and} \bibinfo{person}{Di Wang}.} \bibinfo{year}{2024}\natexlab{}.
\newblock \showarticletitle{MoRAL: MoE Augmented LoRA for LLMs' Lifelong Learning}.
\newblock \bibinfo{journal}{\emph{arXiv preprint arXiv:2402.11260}} (\bibinfo{year}{2024}).
\newblock


\bibitem[Zellers et~al\mbox{.}(2018)]%
        {zellers2018swag}
\bibfield{author}{\bibinfo{person}{Rowan Zellers}, \bibinfo{person}{Yonatan Bisk}, \bibinfo{person}{Roy Schwartz}, {and} \bibinfo{person}{Yejin Choi}.} \bibinfo{year}{2018}\natexlab{}.
\newblock \showarticletitle{Swag: A large-scale adversarial dataset for grounded commonsense inference}.
\newblock \bibinfo{journal}{\emph{arXiv preprint arXiv:1808.05326}} (\bibinfo{year}{2018}).
\newblock


\bibitem[Zhang et~al\mbox{.}(2023)]%
        {zhang2023adalora}
\bibfield{author}{\bibinfo{person}{Qingru Zhang}, \bibinfo{person}{Minshuo Chen}, \bibinfo{person}{Alexander Bukharin}, \bibinfo{person}{Nikos Karampatziakis}, \bibinfo{person}{Pengcheng He}, \bibinfo{person}{Yu Cheng}, \bibinfo{person}{Weizhu Chen}, {and} \bibinfo{person}{Tuo Zhao}.} \bibinfo{year}{2023}\natexlab{}.
\newblock \showarticletitle{Adalora: Adaptive budget allocation for parameter-efficient fine-tuning}.
\newblock \bibinfo{journal}{\emph{arXiv preprint arXiv:2303.10512}} (\bibinfo{year}{2023}).
\newblock


\bibitem[Zhao et~al\mbox{.}(2024)]%
        {zhao2024mosld}
\bibfield{author}{\bibinfo{person}{Lulu Zhao}, \bibinfo{person}{Weihao Zeng}, \bibinfo{person}{Xiaofeng Shi}, {and} \bibinfo{person}{Hua Zhou}.} \bibinfo{year}{2024}\natexlab{}.
\newblock \showarticletitle{MoSLD: An Extremely Parameter-Efficient Mixture-of-Shared LoRAs for Multi-Task Learning}.
\newblock \bibinfo{journal}{\emph{arXiv preprint arXiv:2412.08946}} (\bibinfo{year}{2024}).
\newblock


\bibitem[Zhu et~al\mbox{.}(2023)]%
        {zhu2023sira}
\bibfield{author}{\bibinfo{person}{Yun Zhu}, \bibinfo{person}{Nevan Wichers}, \bibinfo{person}{Chu-Cheng Lin}, \bibinfo{person}{Xinyi Wang}, \bibinfo{person}{Tianlong Chen}, \bibinfo{person}{Lei Shu}, \bibinfo{person}{Han Lu}, \bibinfo{person}{Canoee Liu}, \bibinfo{person}{Liangchen Luo}, \bibinfo{person}{Jindong Chen}, {et~al\mbox{.}}} \bibinfo{year}{2023}\natexlab{}.
\newblock \showarticletitle{Sira: Sparse mixture of low rank adaptation}.
\newblock \bibinfo{journal}{\emph{arXiv preprint arXiv:2311.09179}} (\bibinfo{year}{2023}).
\newblock


\end{thebibliography}

\end{document}